\RequirePackage{amsmath}
\RequirePackage{fix-cm}
\RequirePackage{cite}

\documentclass{article}
\usepackage{lineno}
\modulolinenumbers[1]
\usepackage{amssymb,amsmath,mathtools}
\usepackage{lmodern}
\usepackage{graphicx}
\usepackage{float}
\usepackage{bm,upgreek}
\usepackage{grffile}
\usepackage{etoolbox}
\usepackage{authblk}
\usepackage{xcolor}
\usepackage{xspace}
\usepackage{siunitx}
\usepackage{comment}
\usepackage{orcidlink}
\usepackage{changepage} 
\PassOptionsToPackage{hyphens}{url}
\usepackage{hyperref}

\let\epsilon\varepsilon

\newcommand{\ie}{{\it i.e.,\ }}

\newcommand{\PMten}{PM$_{10}$\xspace}
\newcommand{\PMtwofive}{PM$_{2.5}$\xspace}
\newcommand{\PMCO}{PM$_{CO}$\xspace}
\newcommand{\SOtwo}{SO$_2$\xspace}
\newcommand{\Othree}{O$_3$\xspace}
\newcommand{\CO}{CO\xspace}
\newcommand{\NOtwo}{NO$_2$\xspace}
\newcommand{\NO}{NO\xspace}
\newcommand{\NOX}{NO$_X$\xspace}

\newcommand{\MER}{{\it{MER}}\xspace}
\newcommand{\CAM}{{\it{CAM}}\xspace}
\newcommand{\SAG}{{\it{SAG}}\xspace}
\newcommand{\SFE}{{\it{SFE}}\xspace}
\newcommand{\UIZ}{{\it{UIZ}}\xspace}
\newcommand{\TLA}{{\it{TLA}}\xspace}
\newcommand{\PED}{{\it{PED}}\xspace}

\begin{document}

\title{Tackling air quality with SAPIENS\\[1ex]
\large SAPIENS: Smart Air Pollution Information Enabling New Solutions
}

\author[1]{Marcella Bona\,\orcidlink{0000-0002-9660-580X}}
\author[1]{Nathan Heatley\,\orcidlink{0000-0003-2204-4779}}
\author[1]{Jia-Chen Hua\,\orcidlink{0000-0002-6591-1417}}
\author[2]{Adriana Lara\,\orcidlink{0000-0002-2498-4441}}
\author[1,4]{Valeria Legaria-Santiago\,\orcidlink{0009-0008-4843-4971}}
\author[3]{Alberto Luviano-Ju\'arez\,\orcidlink{0000-0001-8790-4165}}
\author[2]{Fernando Moreno-G\'omez}
\author[1]{Jocelyn Richardson\,\orcidlink{0009-0001-4147-7011}}
\author[2]{Natan Vilchis}
\author[1]{Xiwen Shirley Zheng}

\affil[1]{Department of Physics and Astronomy, Queen Mary University of London, United Kingdom}
\affil[2]{ESFM, Instituto Polit\'ecnico Nacional, Mexico City, Mexico}
\affil[3]{UPIITA, Instituto Polit\'ecnico Nacional, Mexico City, Mexico}
\affil[4]{CIC, Instituto Polit\'ecnico Nacional, Mexico City, Mexico}

\date{\vspace{-5ex}}

\clearpage
\maketitle

\begin{abstract}
Air pollution is a chronic problem in large cities worldwide and awareness is rising as the long-term health implications become clearer. Vehicular traffic has been identified as a major contributor to poor air quality. In a lot of cities the publicly available air quality measurements and forecasts are coarse-grained both in space and time. However, in general, real-time traffic intensity data is openly available in various forms and is fine-grained. In this paper, we present an in-depth study of pollution sensor measurements combined with traffic data from Mexico City. We analyse and model the relationship between traffic intensity and air quality with the aim to provide hyper-local, dynamic air quality forecasts. We developed an innovative method to represent traffic intensities by transforming simple colour-coded traffic maps into concentric ring-based descriptions, enabling improved characterisation of traffic conditions. Using Partial Least Squares Regression, we predict pollution levels based on these newly defined traffic intensities. The model was optimised with various training samples to achieve the best predictive performance and gain insights into the relationship between pollutants and traffic. The workflow we have designed is straightforward and adaptable to other contexts, like other cities beyond the specifics of our dataset.

\end{abstract}

\newpage

\section{Introduction}
\label{sec:intro}
Air pollution is a chronic problem in large cities all over the world, and one of the leading causes of disease and premature death~\cite{Trasande,Raaschou}. Avoiding exposure to air pollutants is especially important for susceptible individuals with chronic cardiovascular or pulmonary disease, children, and the elderly~\cite{Zuurbier}.
Governments and global activist movements are raising awareness and promoting actions to tackle poor quality air, which regularly reaches illegal and unsafe levels worldwide.
There are many contributing factors, including traffic, geography, city planning,
industry, commerce and domestic emissions. In particular, private, public and
commercial traffic have been identified as major contributors~\cite{Ghosh,Bel,Hilpert}, affecting citizens’ health, the economy, and tourism.
However, mitigations are possible. Individuals who commute to work in personal
vehicles or public transportation receive a substantial portion of their daily dose of air pollution during commuting activities~\cite{Zuurbier}. Personal exposure to ambient air pollution can therefore be reduced on high air-pollution days by better choice of travel routes~\cite{Laumbach}. One study in London has found that using side streets can cut air pollution exposure by half relative to using the most polluted routes~\cite{Guardian}. Public transport policy may influence patterns of pollution~\cite{Bel}, while studies in Mexico~\cite{Chakraborti_Voorheis_2025} show that the effect of social inequalities results in an almost four-fold annual increase of pollution concentrations, which can lead to a reduction in public health for vulnerable populations.

Awareness of air pollution levels has been increased by academic studies, although many cover relatively short periods of time, forcing researchers to use projections or incomplete data.
There are also a growing number of public air quality alert systems. In London there is now hyper-local air quality information driven by a high-quality and dense sensor network. Such high-quality information, with resolution at the level of individual streets, enables citizens to make informed choices about travel routes in order to minimise their exposure to pollutants generated by traffic. Further, since air quality has strong daily cycles~\cite{Chen}, information which is updated regularly enables the best routes to be chosen according to the time of day.
The availability of specific and local information is also important for policy makers.
High-resolution spatio-temporal measurements enable local governments to identify
problem areas for air quality, both by location and also by time of day. This in turn enables urban planning for the mitigation of these issues, including better informed development and traffic management.
In total, the benefits of reduced exposure to air pollution include improved quality of life, including citizens’ mental and physical health and reduced health care costs. 

In this paper, we focus on data from Mexico City to develop a proof-of-concept air quality prediction system. This system aims to provide hyper-local, dynamic air quality forecasts using existing open data sources. In particular, openly available traffic intensity data are leveraged to address the low density of pollution sensors.
Currently, citizens of Mexico City have access to only a single air pollution measurement per county, available via a smartphone app~\cite{airecdmx}. Although the local government may consider expanding the sensor network in the future, the current distribution leaves large areas of the city without specific information on local air pollution levels.

The SAPIENS (Smart Air Pollution Information Enabling New
Solutions)~\footnote{\url{https://sapiens.qmul.ac.uk/}}
project has begun building the SAPIENS database, which contains information on traffic levels as well as measurements from air pollution sensors across Mexico City.
The goal is to model the relationship between traffic and pollution and to generate pollution predictions based on traffic information.

At the level of input data, the most closely related recent work is Ref.~\cite{zalakeviciute2020traffic}, where Google Maps traffic data is used to predict~\PMtwofive. The authors use a decision-tree machine-learning algorithm and the results show that traffic-based prediction outperforms interpolation and that adding the time of day further increases accuracy on average. However, it is not clear how the Google Maps traffic data were quantified in their study, whereas in this paper we develop a new strategy for this purpose.
At the level of data analysis, Ref.~\cite{zhang2012real} provides a useful historical perspective of the more general problem of real-time air quality forecasting, including three-dimensional modelling and the complex interplay of meteorology, emissions, and chemistry, from global to urban scales. Their conclusions support the use of traffic information and statistical methods, which is the starting point for this paper, and suggest the inclusion of physical models and meteorological data which will be part of the future developments of the SAPIENS analysis. Ref.~\cite{eslami2019data} focuses on day-ahead prediction of hourly ozone levels and proposes using an ensemble of machine learning models rather than single models to improve performance.

In addition to the link between traffic emissions and air pollutants, the impact of traffic speed on air pollution levels is also considered vital for investigating air quality improvements. A review~\cite{emission_speed} found that most studies reported that a reduction in speed leads to lower emissions of \NOX and particulate matter, although the effects varied greatly across studies, depending on a variety of factors (the range of speed limit reductions, the type of pollutant, accompanying interventions, the methodologies, etc.).
It has also been observed that driving involving frequent acceleration, deceleration, stops, and starts increases air pollution from both exhaust emissions and non-exhaust sources such as brake pad and tyre wear, with the latter accounting for over 75\% of road transport particulate emissions~\cite{Speed}.
With SAPIENS, we analyse the relationship between traffic and pollution, taking into account the different traffic conditions (free-flowing versus stop-start) as they can be derived from Google Maps congestion levels.

\section{Data Overview and Analysis}
\label{sec:data}

As the SAPIENS project aims at predicting pollution levels from traffic information,
we have developed a complete workflow from input data collection and analysis to model assessment and result validation. The SAPIENS database
contains information and measurements on traffic levels and air pollution. We
used a Partial Least Squares regression (PLSR) to infer pollution levels from traffic
intensities.

\begin{figure}[htb!]
  \centering
    \vspace*{-0.1cm}
  \includegraphics[width=0.70\textwidth]{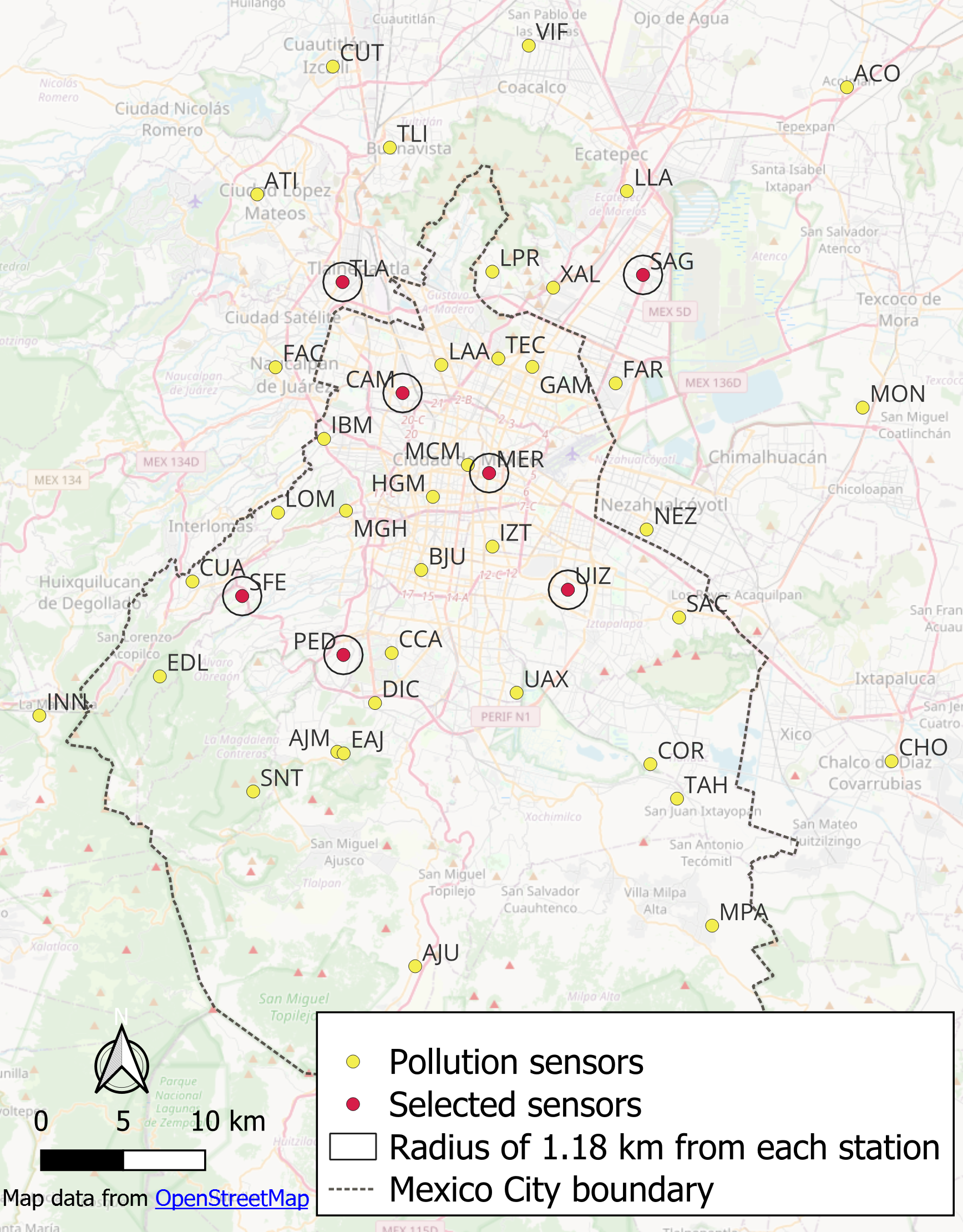}
      \vspace*{-0.3cm}
  \caption{Geographical locations of the 44 pollution sensors available in Mexico City. The red dots represent sensors whose measurements we use in this paper and the circles around them represent the 1.18km-radius areas we consider for traffic estimation.}
  \label{fig:sensorsmap}
\end{figure}

The air pollution measurements consist of data coming from air pollution sensors placed in 44 locations across the Mexico City metropolitan area as seen in Fig.~\ref{fig:sensorsmap}. These data are obtained from the Mexico City atmospheric monitoring directorate~\cite{airecdmx_values}. The time period we consider in this study is 14$^{th}$ December 2020 to 1$^{st}$ April 2021.

We performed an analysis of these sensor data to identify their usability. There are missing (null) values in the data for many stations, either for specific pollutants or for all of them. We selected the seven stations with the highest percentage of complete sets of measurements in the considered period (labelled as \CAM, \MER, \TLA, \SAG, \SFE, \UIZ, and \PED, and indicated with red dots in Fig.~\ref{fig:sensorsmap}). We define a complete set as having a full measurement for all the nine pollutants at a given time.
The maximum number of possible sets for this time period is 2,610 corresponding to 109 days with 24 records per day. With respect to this, the percentage of complete sets ranges between $88\%$ and $45\%$ for the top seven stations.

Traffic patterns are obtained by looking at live Google traffic maps, centred on the positions of the air pollution sensors. These images were processed to calculate representative values, which we define as "traffic intensities". This data is collected every hour in this time period, details of the traffic data is found in Sec.~\ref{subsec:GoogleTraffic}.

Our preliminary data analysis showed that most of the outliers and irregularity in the traffic happen during holidays and nights. Therefore, for simplicity and to obtain a cleaner sample, we filtered out the weekends and main holidays, and selected only the core hours of working days (Mon-Fri 7am to 10pm), during which the traffic activities are intense and have clearer patterns.
After this selection, the percentage of data used ranges between $34\%$ and $23\%$ across the stations.

\subsection{Air Pollution Analysis}
\label{subsec:PollutantsConcentration} 
We analysed 
hourly measurements of nine pollutants, \PMten ($\mu g/m^3$), \PMtwofive ($\mu g/m^3$), \PMCO ($\mu g/m^3$), 
\SOtwo (ppb), \Othree (ppb), \CO (ppm), \NOtwo (ppb), \NO (ppb), and \NOX (ppb).

\begin{figure}[htb!]
  \centering
  \includegraphics[width=0.95\textwidth]{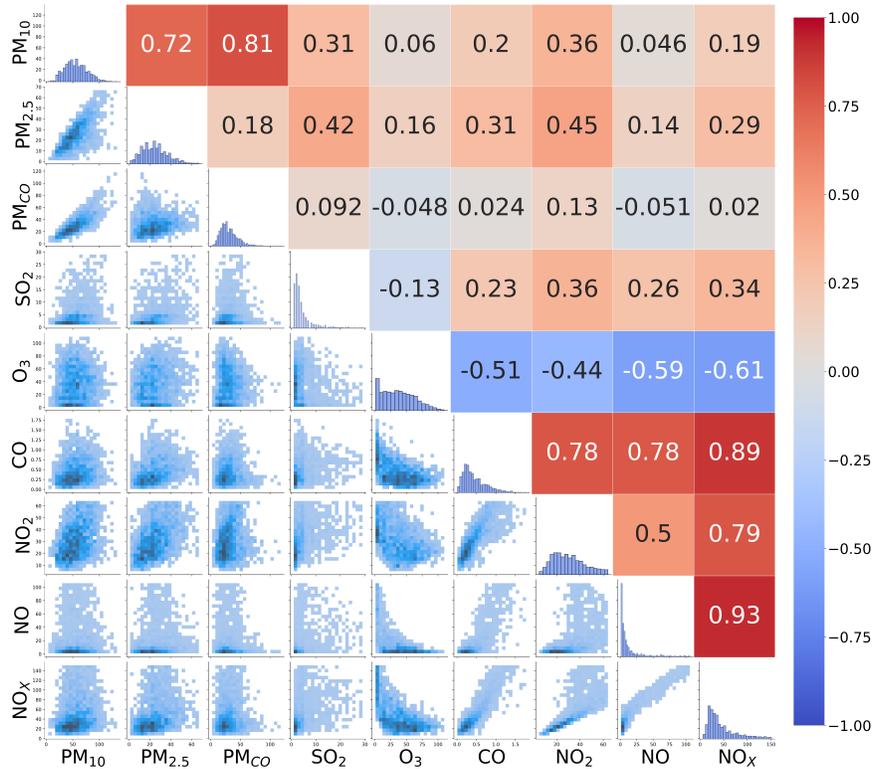}
\caption{Correlations between the measurements of the nine pollutants as evaluated from the \CAM station data.}
\label{fig:pollutantscorrelations}
\end{figure}

Fig.~\ref{fig:pollutantscorrelations} shows the linear correlations between the nine pollutants calculated using the \CAM station. Colours towards the red indicate a strong positive correlation, while colours towards the blue indicate a strong negative correlation.
Two groups of correlated pollutants are identified. The upper-left corner of the matrix contains pollutants \PMten, \PMtwofive and \PMCO which are correlated as expected as they are all measuring particulate matter, \PMten being the sum of \PMtwofive and \PMCO. The lower-right corner of the matrix contains the four pollutants \CO, \NOtwo, \NO, and \NOX which are also positively correlated: in particular, \NO and \NOtwo are subsets of \NOX. Ozone is anti-correlated with the nitrous oxide pollutants, while the sulfur dioxide (\SOtwo) is uncorrelated with all the others. 

\begin{figure}[htb!]
  \centering
\hspace*{-1.8cm}
  \includegraphics[width=0.37\textwidth]{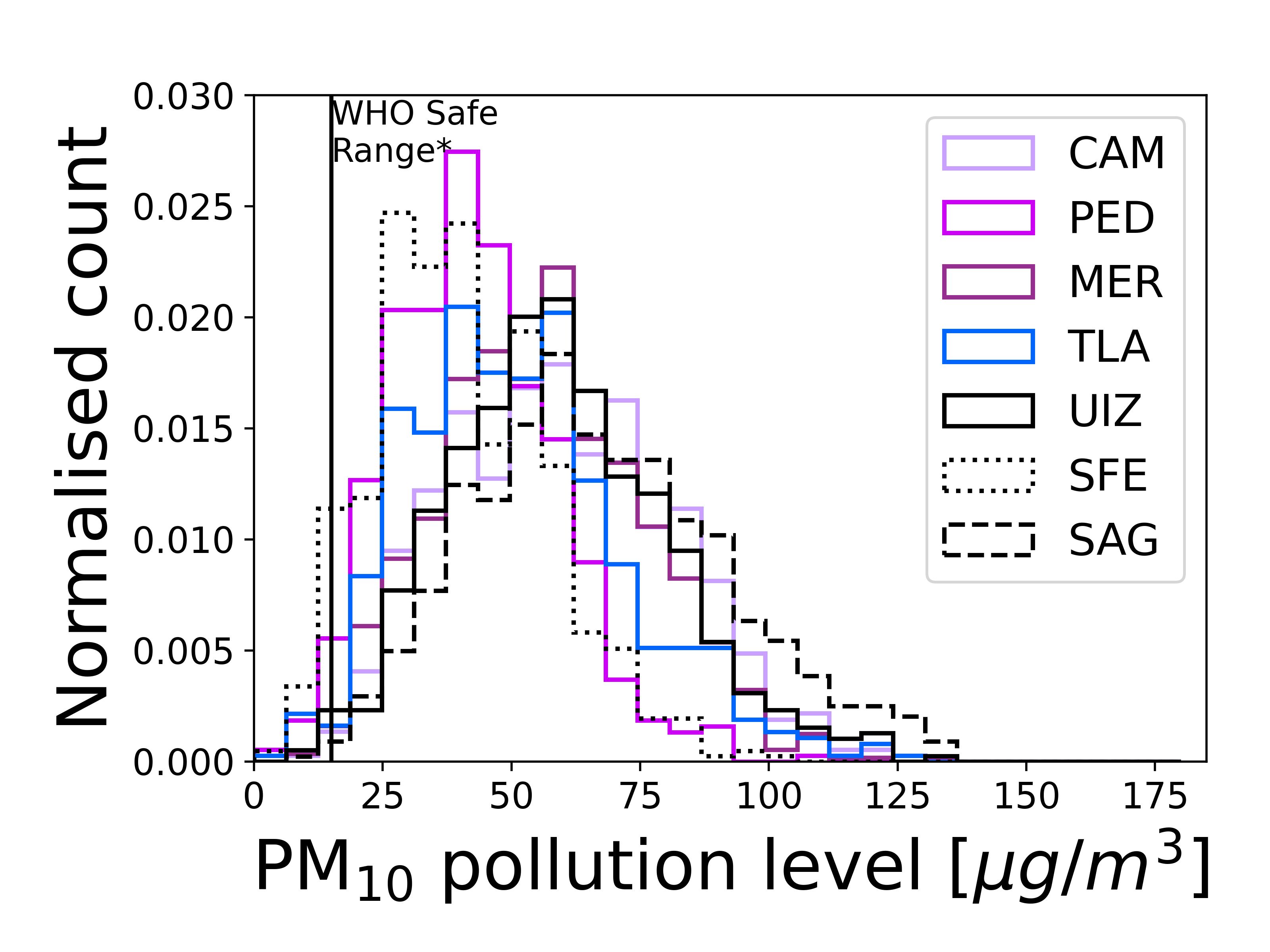}
  \includegraphics[width=0.37\textwidth]{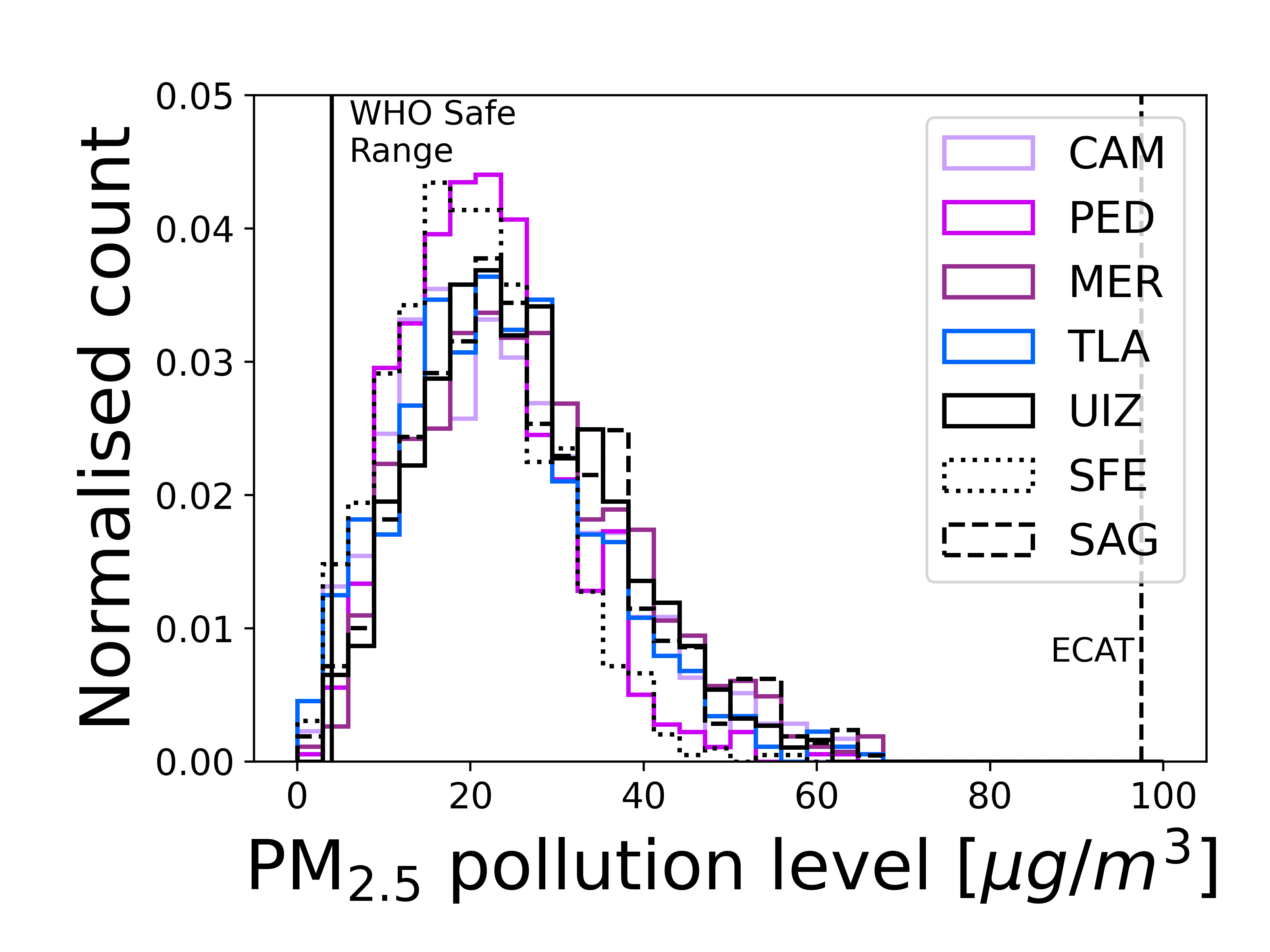}
   \includegraphics[width=0.37\textwidth]{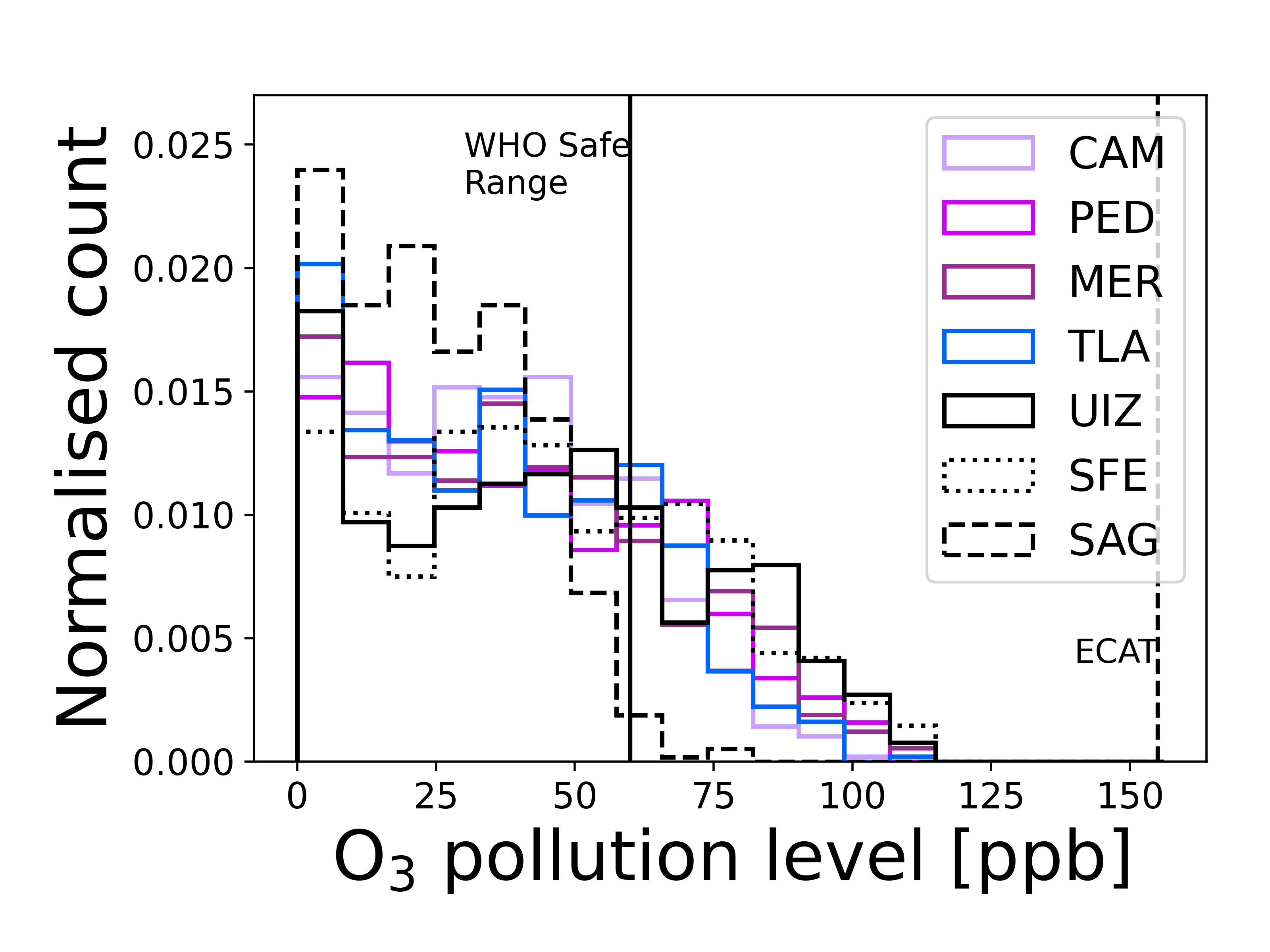}  
  \hspace*{-1.3cm}
\caption{Distributions of the measurements of \PMten, \PMtwofive and \Othree in the seven stations considered. The vertical lines (when present) correspond to the annual exposure ranges (solid line), as calculated by the WHO~\cite{WHO_exposurelevels} and the thresholds (dashed line) linked to the Mexico City environmental atmospheric contingency~\cite{CDMXthreshold}.}
\label{fig:pollutants}
\end{figure}

Fig.~\ref{fig:pollutants} shows the distributions of the measurements of \PMten, \PMtwofive and \Othree in all the seven stations considered, compared with the WHO exposure ranges~\cite{WHO_exposurelevels} and the thresholds used in Mexico City to trigger the environmental atmospheric contingency restrictions. The environmental atmospheric contingency is declared when \Othree, \PMten, or \PMtwofive levels exceed the given threshold. In this case, a public policy is activated which restricts vehicle traffic based on license plate numbers. The policy is deactivated when the levels of the pollutants that trigger it fall below the threshold. A newspaper source~\cite{CDMXthreshold} quoted the \Othree threshold at 155 ppb, \PMten threshold at 214 $\mu g/m^3$, and \PMtwofive threshold at 97.4 $\mu g/m^3$.

\begin{figure}[htb!]
  \centering
\hspace{-1.3cm}
  \includegraphics[width=0.37\textwidth]{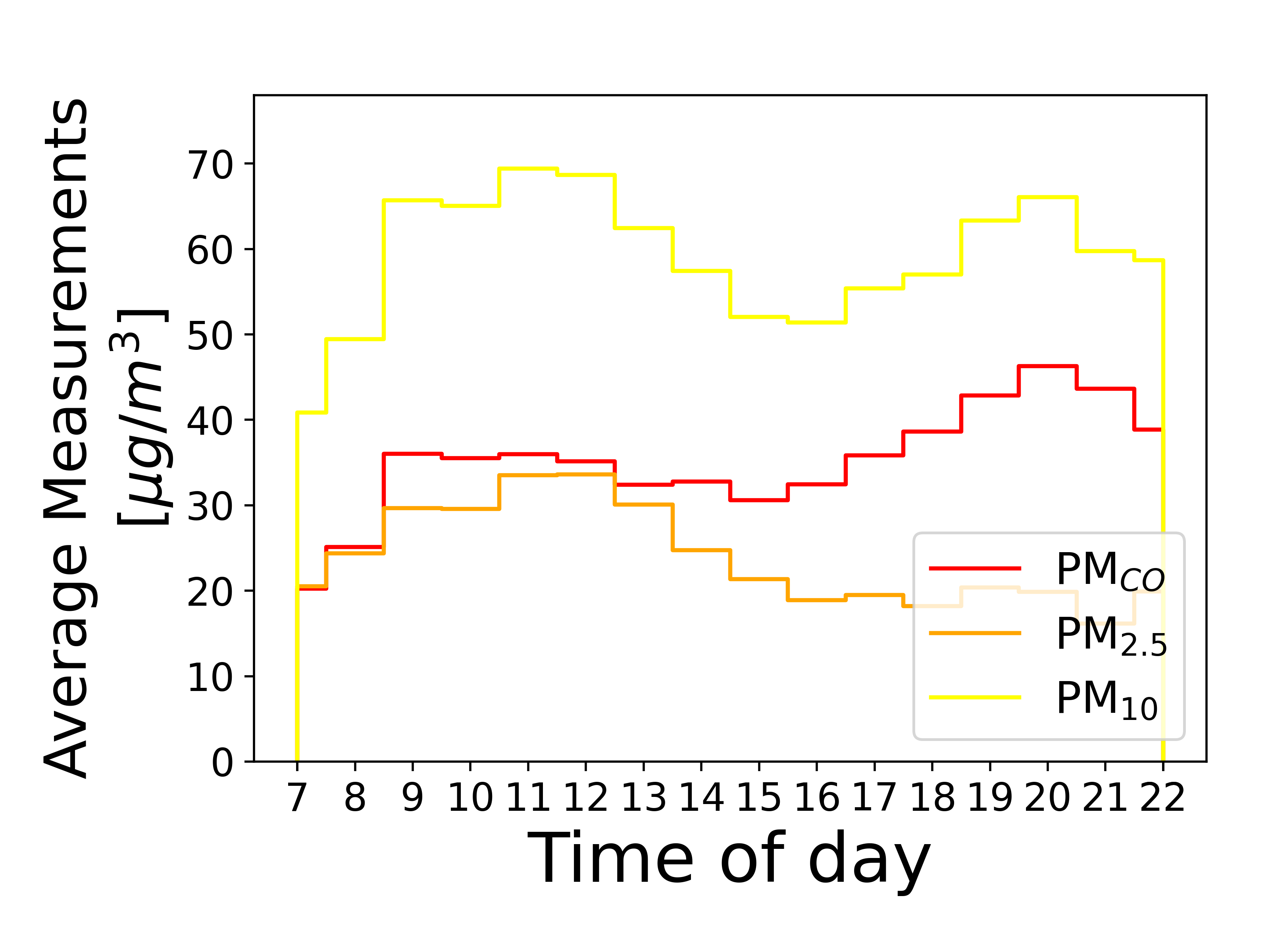}
\hspace{-0.3cm}
  \includegraphics[width=0.37\textwidth]{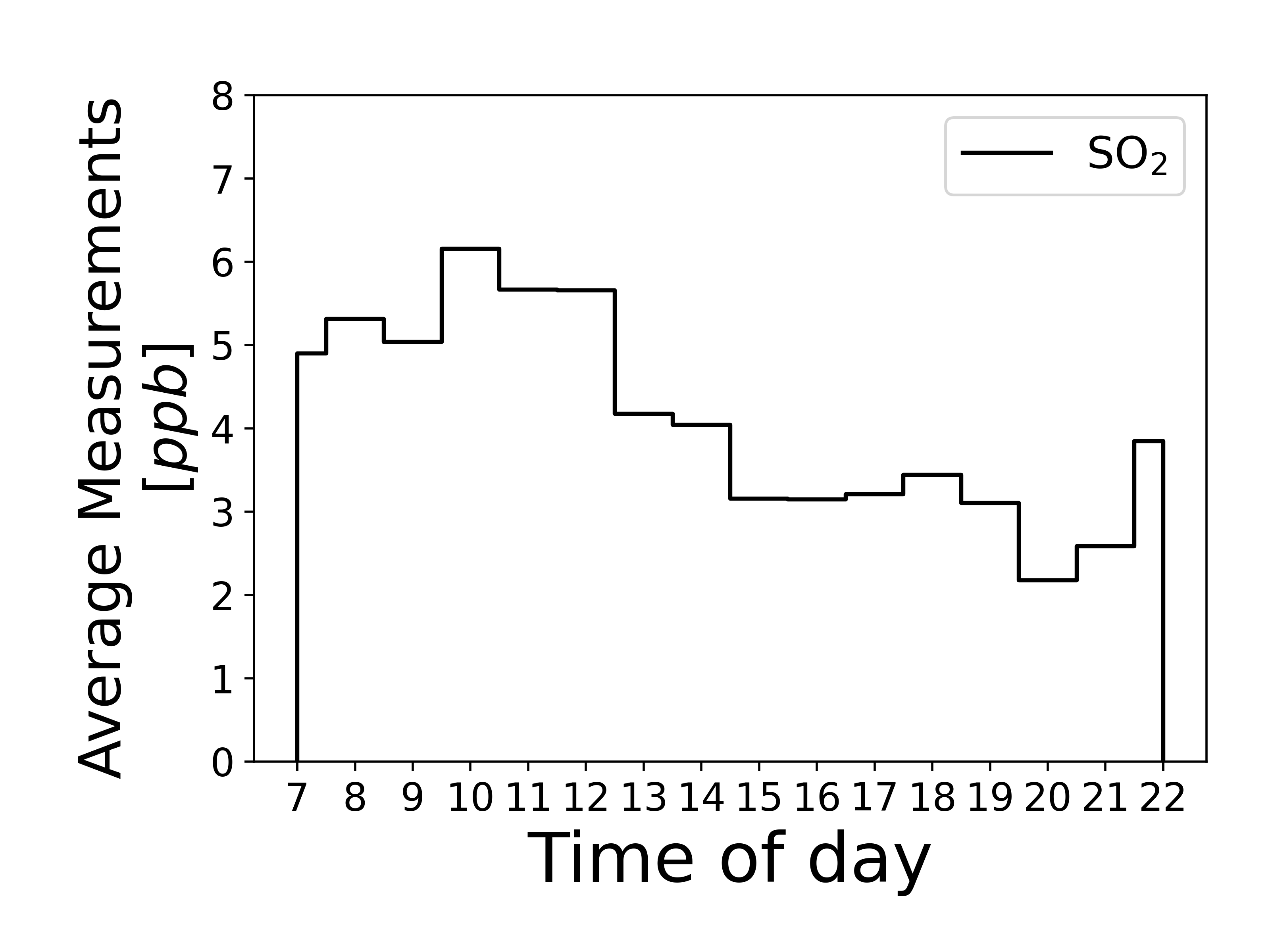}
\hspace{-0.3cm}
  \includegraphics[width=0.37\textwidth]{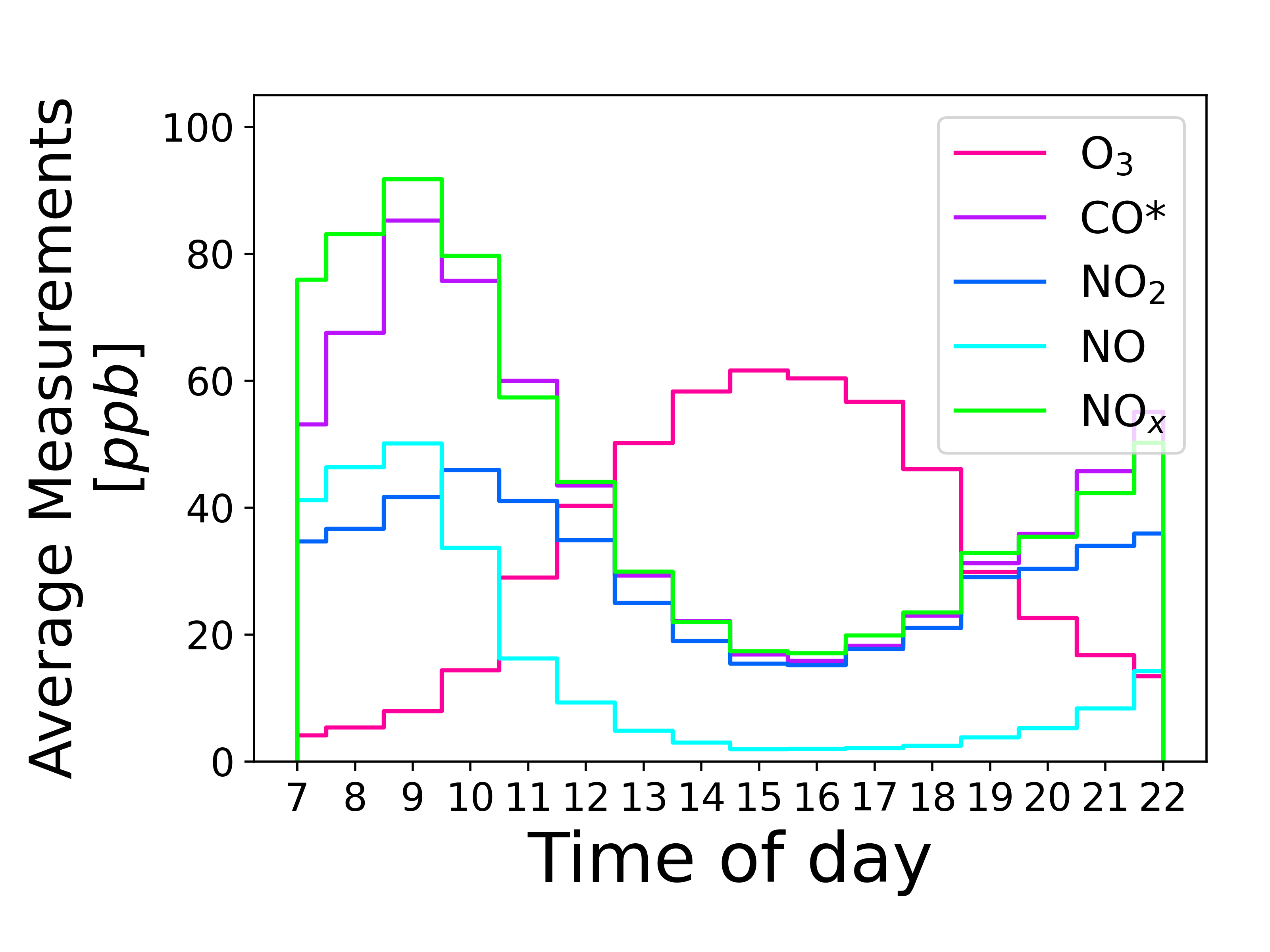}
  \hspace{-1.3cm}
\caption{Distributions of the measurements of the nine pollutants as a function of the time of the day as measured by the \CAM station. Pollutants are grouped based on their (anti-)correlations. 
*Note that CO is measured in ppm, but to compare shapes of distributions, we scaled the measurements by a factor of 100.}
\label{fig:pollutantshourly}
\end{figure}

Fig.~\ref{fig:pollutantshourly} shows the behaviour of the pollutant average values as a function of the time of the day for the \CAM station. 
The measurements of the nitrous oxide pollutants are higher in the morning compared to the rest of the day, while \Othree has a large peak in the middle of the day. 
This is consistent with the \Othree formation process. \Othree and \NO are formed when ultraviolet radiation from the sun provokes chemical reactions between \NOX and volatile organic compounds~\cite{ozonemalaysia,ozoneprecursors,ozonereactions}. As temperature and sunlight increase from morning to midday, \Othree levels increase to reach a maximum at noon, while \NOX and \CO are at a minimum. After this point, there is a decline in \Othree as sunlight decreases, which explains the negative correlation between \Othree and the nitrous oxide group of pollutants seen in Fig.~\ref{fig:pollutantscorrelations}. 
The particulate matter group shows higher pollutant measurements in the morning and evening, coinciding with the rush hour traffic periods.

The hourly measurements of these pollutants are used later to train the PLSR, as they represent the primary target of our predictive model.

\subsection{Traffic Analysis}
\label{subsec:GoogleTraffic}
Google Maps images were used to define a traffic intensity figure of merit. Google Maps defines four modes of traffic identified by four colours on the maps ~\cite{Mapscolours}:
\begin{itemize}
  \setlength{\itemsep}{0cm}
  \setlength{\parskip}{0cm}
    \item Green: No traffic delays.
    \item Orange: Medium amount of traffic.
    \item Red: Traffic delays.
    \item Dark red: Major traffic delays (traffic jams)
\end{itemize}

We use these colours to define the inputs to our model as to represent different levels of traffic contributions.

The time resolution of pollutant concentration recordings is one hour, meanwhile Google updates the traffic information on Google Maps every 10 to 15 minutes. In order to match the time measurements of the pollutants and the traffic, an effective time resolution of one hour was taken for the traffic as well.

Since the pollutant concentration is a physical density (\ie a scalar field), the traffic intensity, which is to be defined and mapped to the pollutant concentration, should also be a physical field.
The resolution of the images considered is HD ($1920\times1080$), corresponding to more than 2 million pixels. We aggregate the pixels to get a measurement of traffic intensity that is suitable to be used as an independent and numerical feature.

Inspired by Fig.~A.5 in Ref.~\cite{hilpert2019new}, we propose a definition of the traffic intensity based on the percentage of the same-coloured pixels in predefined discretised areas around the sensor location.
\begin{figure}[htb!]
	\begin{center}
        \includegraphics[width=0.99\textwidth]{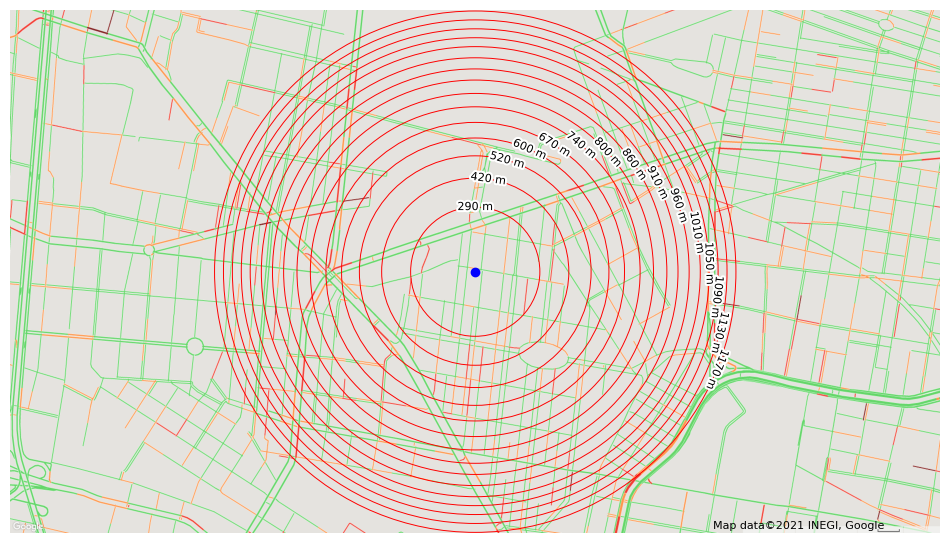}
	\end{center}
    \caption{The road layout around the \CAM pollution sensor station. Here the different traffic levels on the roads are denoted by their Google colours. The 15 rings centred on the pollution sensor position and used for defining traffic intensities are drawn.}
    \label{fig:Station15Rings}
\end{figure}
 
Fig.~\ref{fig:Station15Rings} shows a snapshot of the traffic distribution around the \CAM sensor station and the areas to be used to aggregate the pixels into 15 rings.
We can consider annuli or further split into angular sectors. The latter discretisation could be made to match the wind directions, which are usually reported in terms of 8 (classic denominations), 12 (as in Ref.~\cite{hilpert2019new}) or 16 directions (as in Ref.~\cite{weatherChannel}~and Ref.~\cite{BBCweather}). This can be used to incorporate meteorological information, which also influences pollutant concentration measurements.

For this initial analysis, we aim at keeping our modelling as simple as possible, and hence we consider only the annular discretisation and we calculate the percentages of coloured pixels in each annular ring.
The total traffic intensity $I$ for each colour $c$ is thus defined as:
\begin{equation}
\label{eq:intensity}
I^c = \frac{1}{N}\sum_i \frac{n_i^c}{t_i}
\end{equation}
where $n_i^c$ is the number of pixels of colour $c$ counted in ring $i$, $t_i$ is the total number of pixels in ring $i$ and $N$ is the total number of rings. Given the above definition, the values of the four colour intensities do not add up to one, as some pixels in the image correspond to non-road areas, such as buildings or other features, hence they contribute to the total pixel normalisation, but not to the traffic pixels.

In this analysis, we consider $15$ rings, such that each has roughly the same number of pixels, as visualised in Fig.~\ref{fig:Station15Rings}. As a consequence, we define four traffic intensities, one for each colour, calculated summing over the 15 rings ($c=4$ and $i=15$ in Fig.~\ref{eq:intensity}).

\begin{figure}[htb!]
	\begin{center}
		\includegraphics[width=0.45\linewidth]{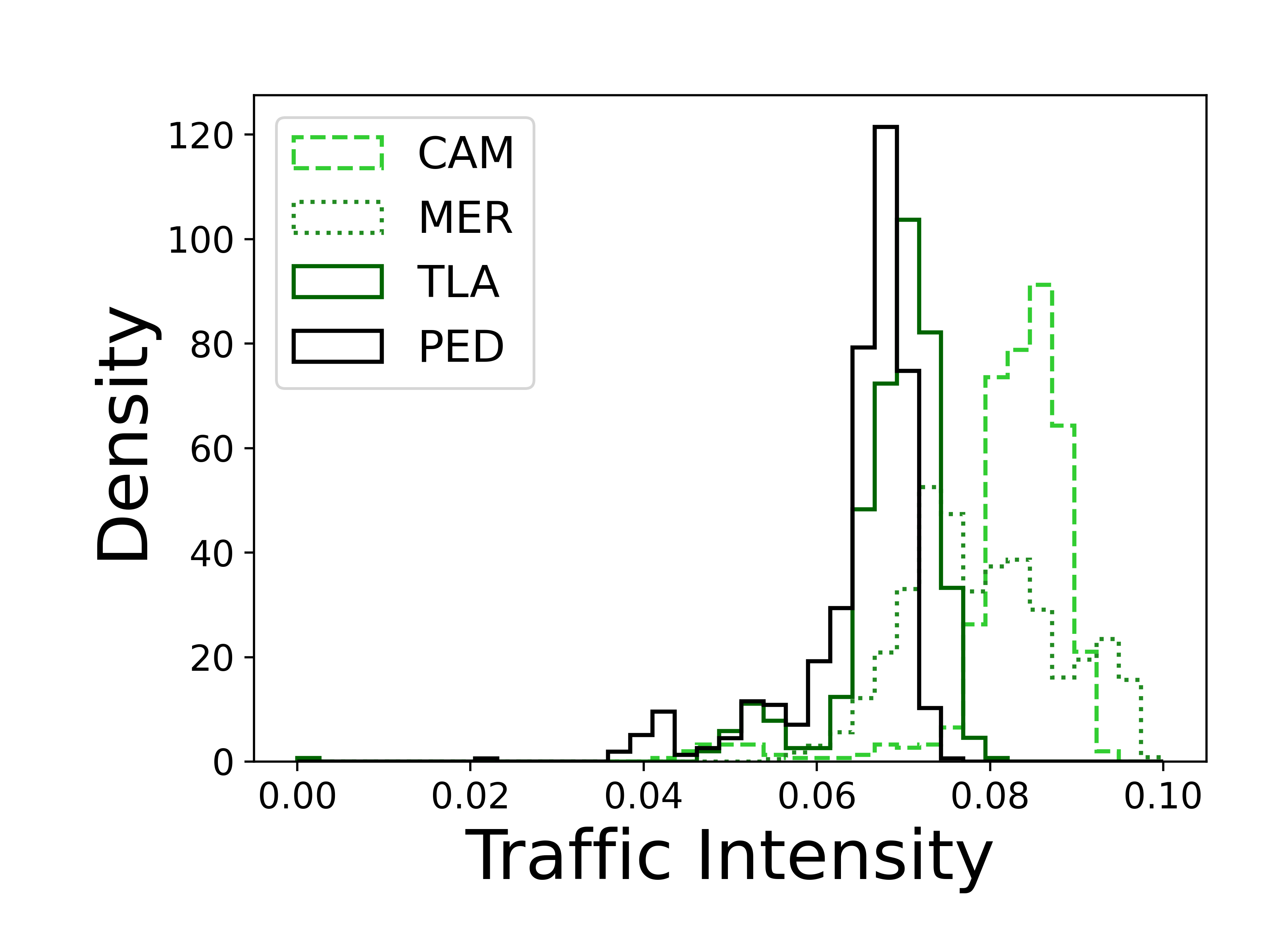}
  		\includegraphics[width=0.45\linewidth]{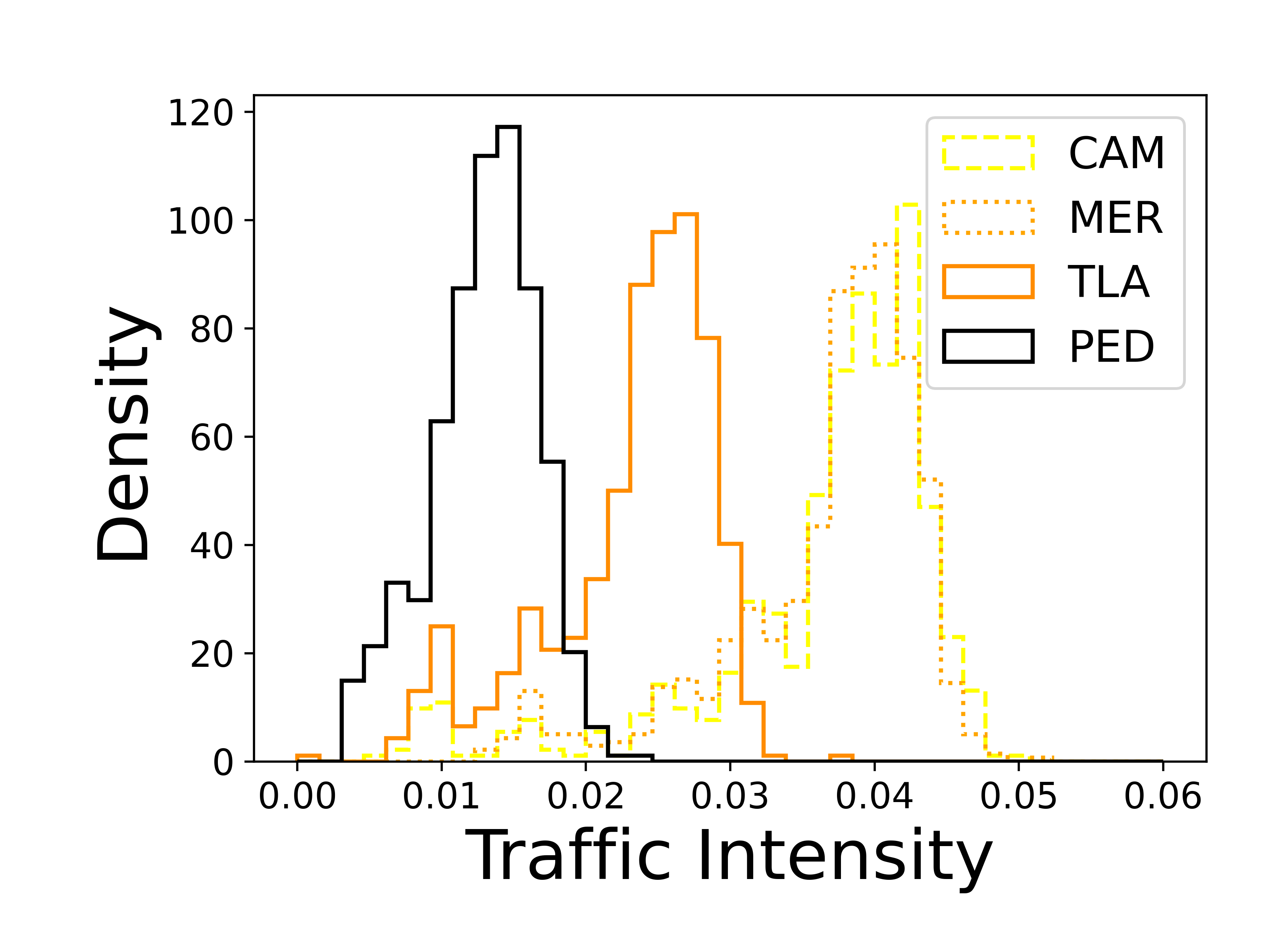}
		\includegraphics[width=0.45\linewidth]{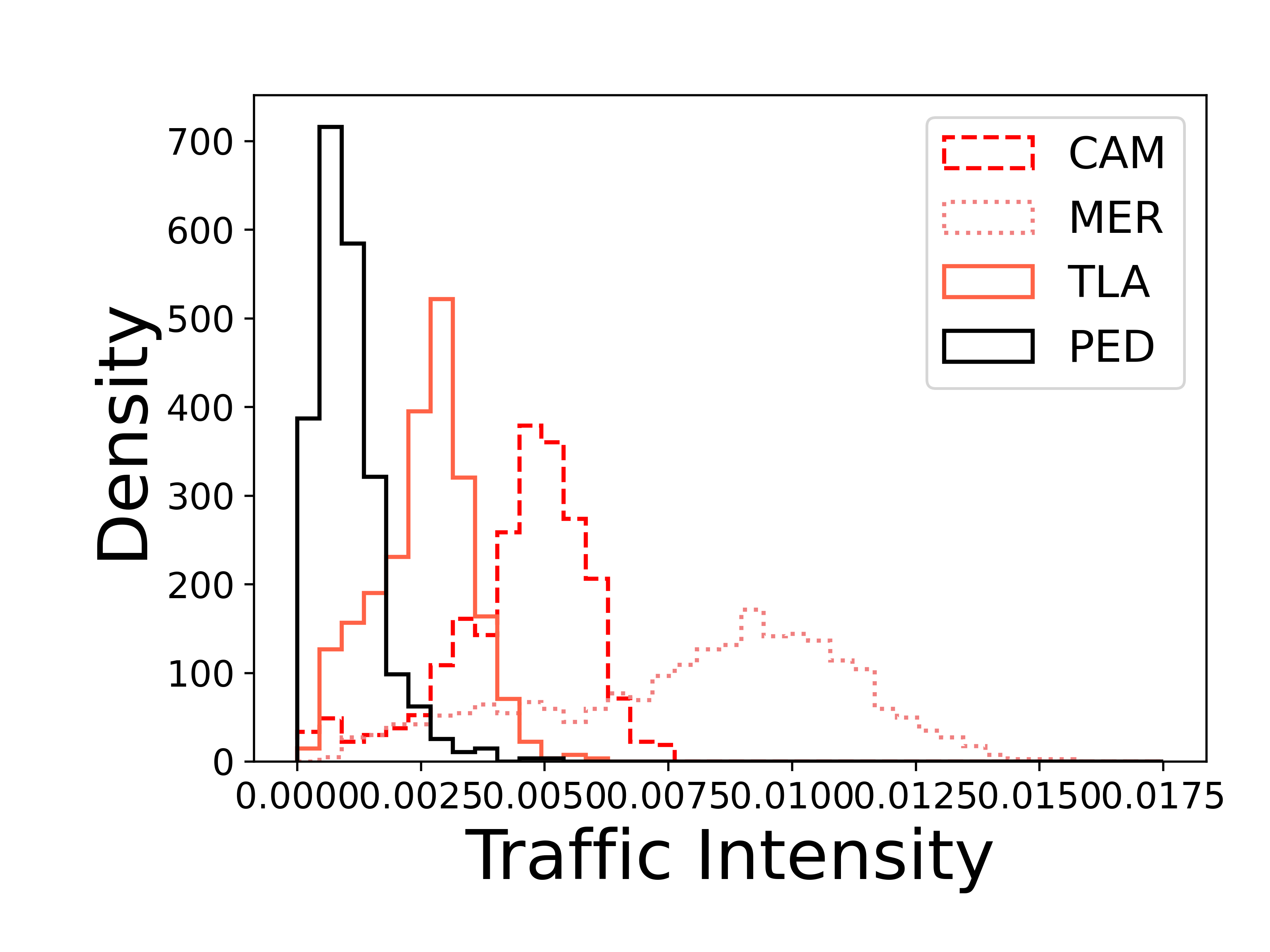}
		\includegraphics[width=0.45\linewidth]{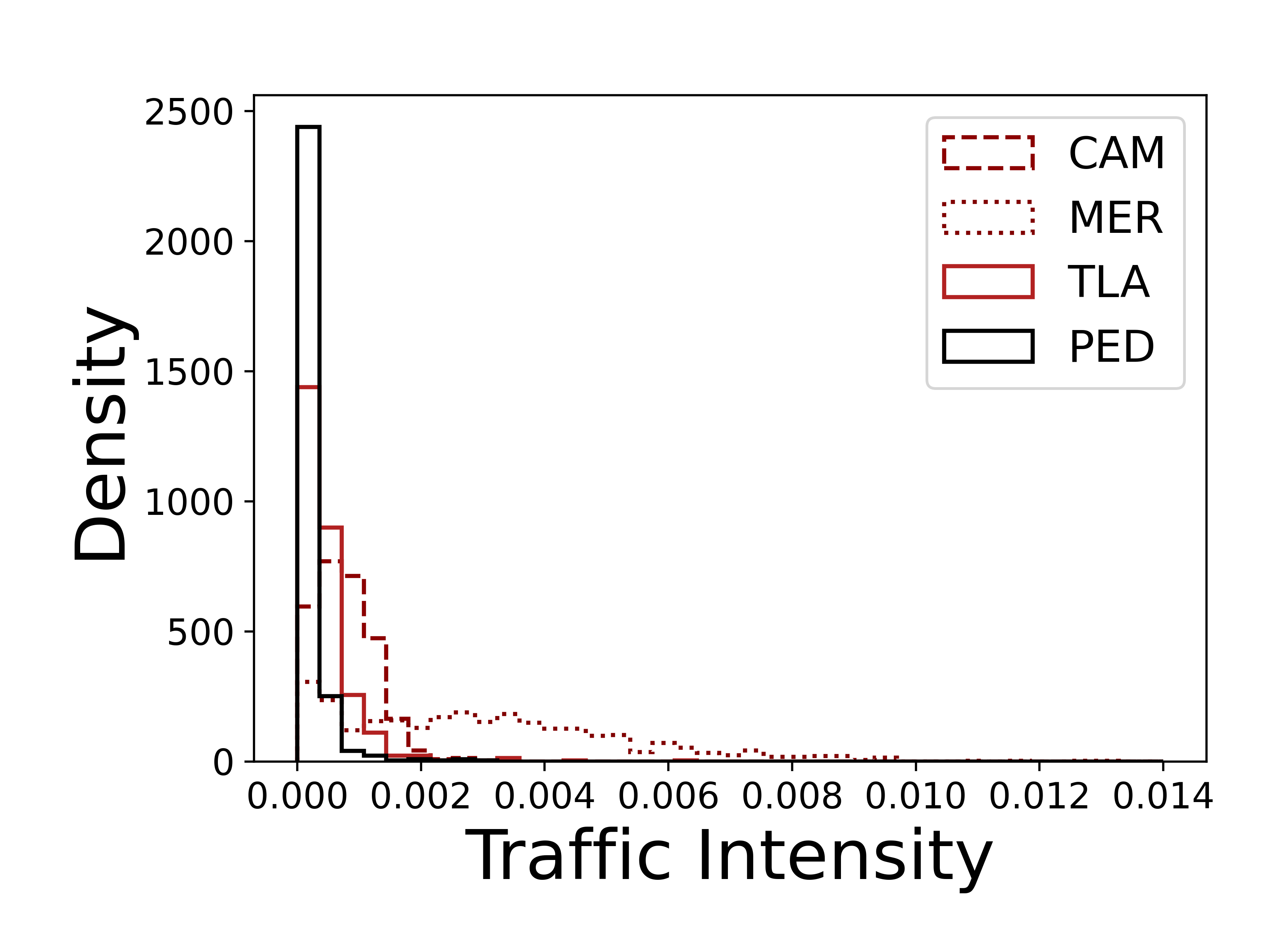}
		\caption{The distributions of the traffic intensities in each colour for the four selected stations.}
		\label{fig:DistributionTrafficByColour}
	\end{center}
\end{figure}

Fig.~\ref{fig:DistributionTrafficByColour} shows the traffic intensity density distributions for the four colours around the four pollution stations \CAM, \MER, \TLA, and \PED, where the first three are selected as benchmark training stations, and the last is used as validation station in the PLSR analysis.

The intensity is highest for green and decreases for subsequent colours, with red/dark red having the lowest averages. 
This is expected as high traffic will occur in localised areas or time periods hence its global intensity will be relatively small with respect to the other colours.
As these distributions are obtained by integrating both in time and rings, certain short periods of the day with anomalous traffic behaviour (i.e. early morning or peak rush hour) may result in low or high tails.
Otherwise, the intensities can be seen to behave in a Gaussian manner.

The areas connected to the four stations show reasonably similar behaviour in the distribution shapes, while the area around the \MER station shows significant discrepancies in the distributions especially of the high traffic colour intensities (red and dark red) with respect to the other stations. This difference could be related to the road topology of the area, as the \MER station is located in the city centre near the central bus terminal, which tends to be congested.
\PED and \TLA have overall lower levels of traffic intensity, compared to the other two stations, for every colour. This can be explained by the distinct road topology, as the two most traffic-heavy roads are on the outer boundary of the area considered. For these stations, the dark red traffic has a very low intensity indicating a road topology that is resistant to congestion, or that the zone is rarely over-burdened by traffic.

\begin{figure}[htb!]
	\begin{center}
		\includegraphics[width=0.45\linewidth]{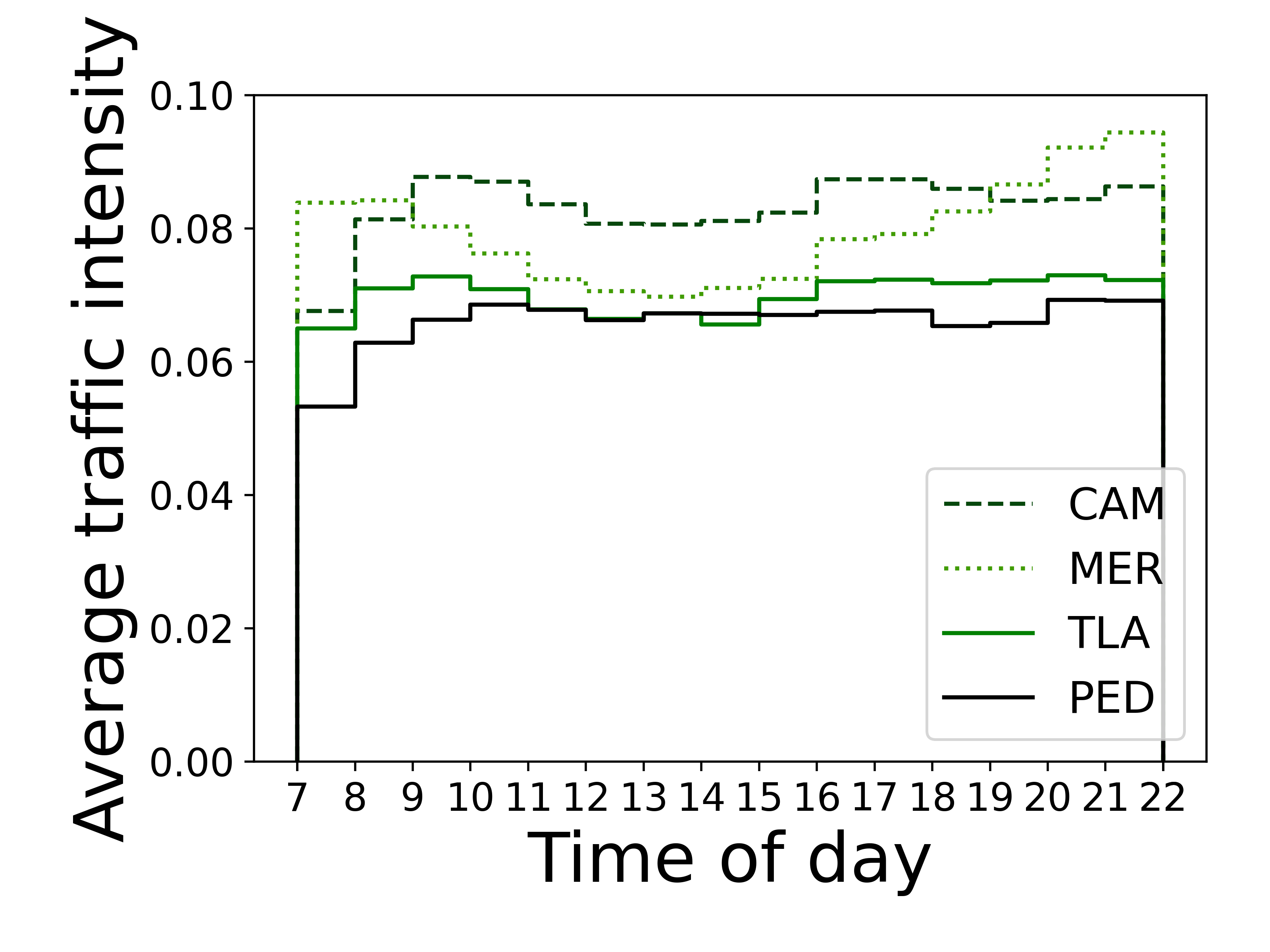}
  		\includegraphics[width=0.45\linewidth]{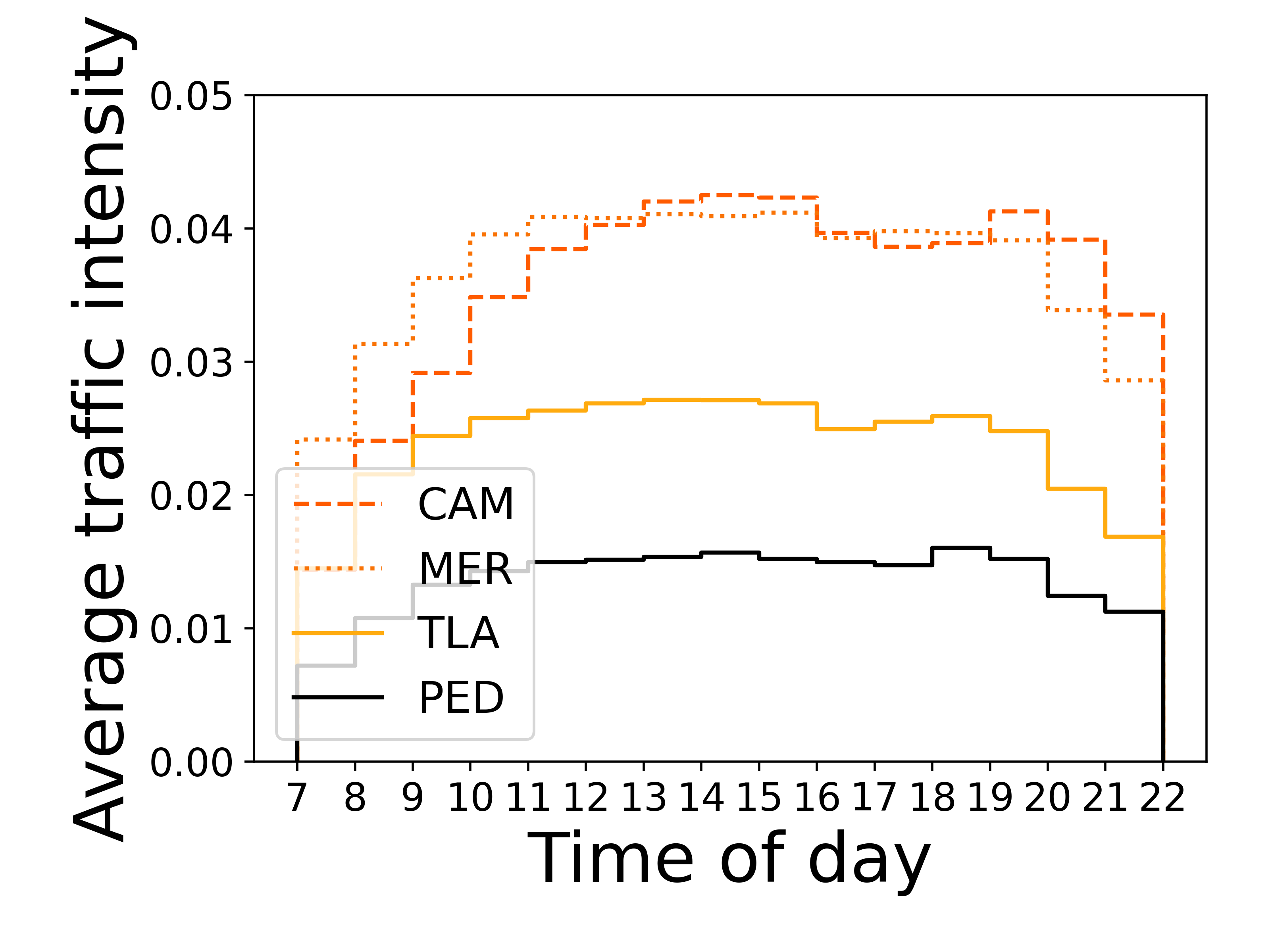}
		\includegraphics[width=0.45\linewidth]{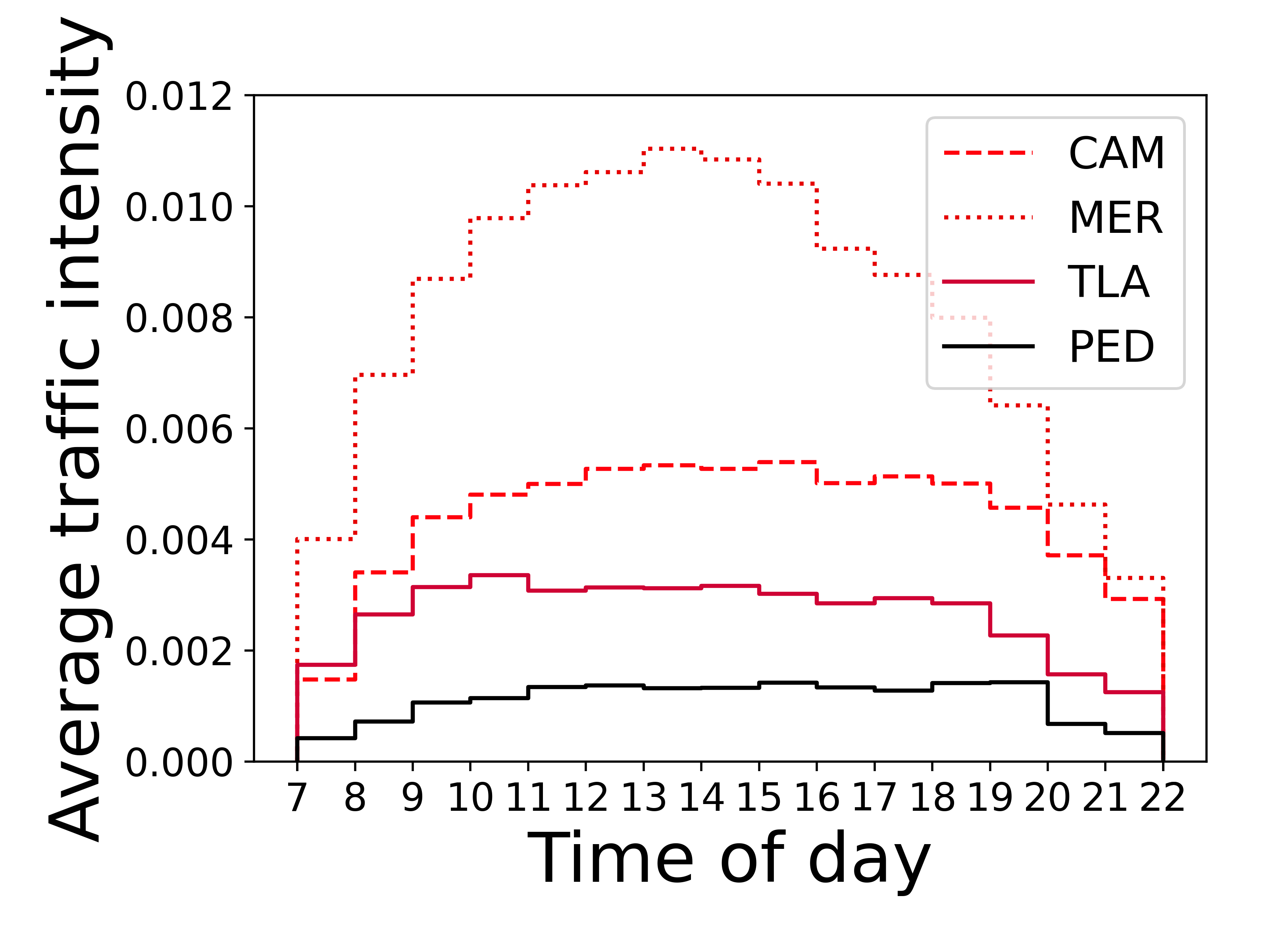}
		\includegraphics[width=0.45\linewidth]{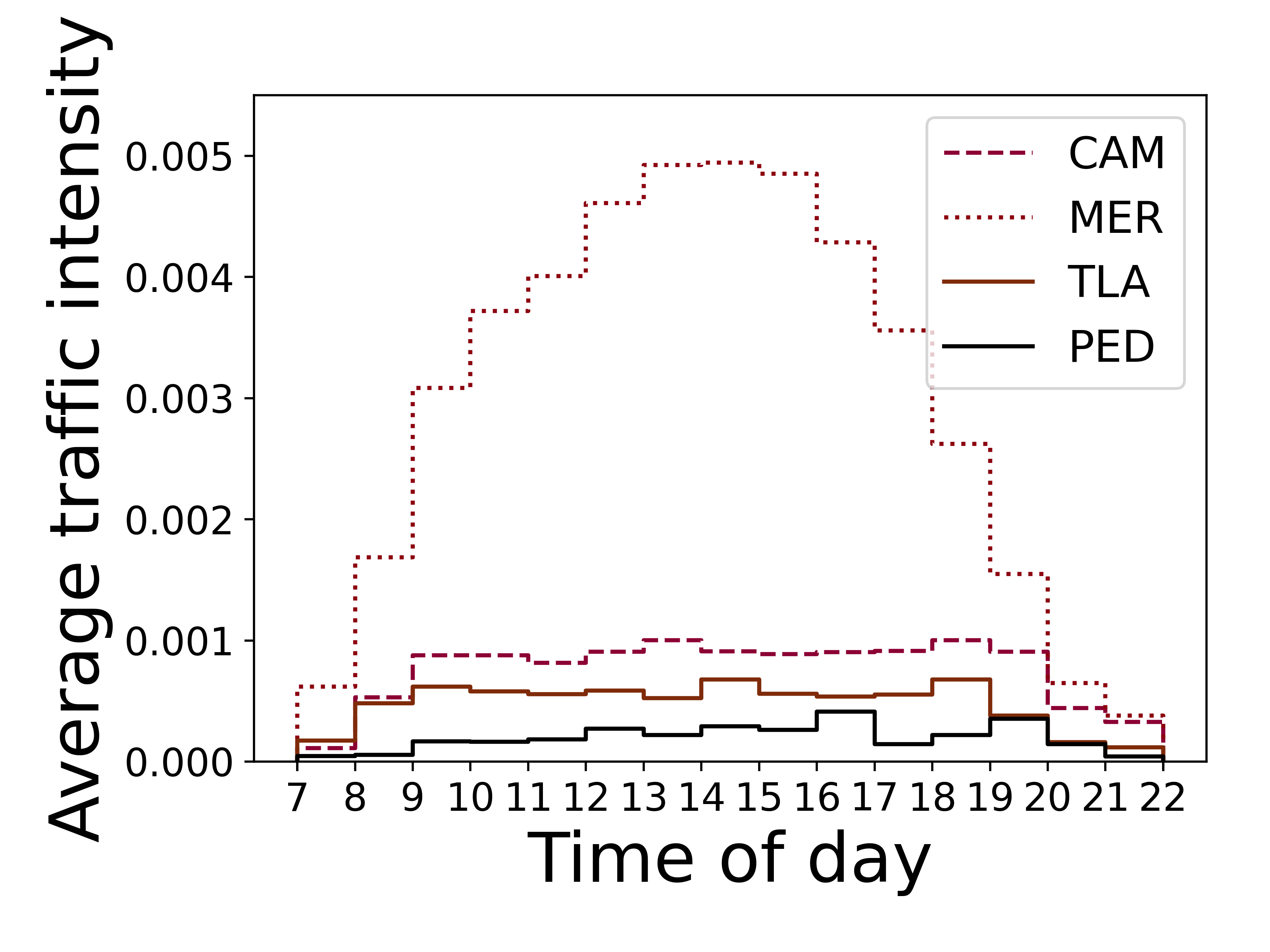}
		\caption{Distributions of the average traffic intensity (i.e. total percentage of coloured pixels within the disc) against the local time for each traffic colour at the four stations \CAM, \MER, \TLA and \PED.
        Note that the values of the four colour intensities do not sum to one, as some pixels in the image correspond to non-road areas, such as buildings or other features. These pixels are included in the total pixel normalisation but do not contribute to the traffic pixels.}
		\label{fig:TrafficTimesbyColour}
	\end{center}
\end{figure}

Fig.~\ref{fig:TrafficTimesbyColour} shows the hourly traffic intensity behaviour by colour across the four selected stations. The green distributions display two slight peaks corresponding to rush hours—one in the morning (around 9–11~am) and another in the evening after 4~pm. These peaks reflect an increase of vehicle presence on the streets, which results in more data being recorded in the traffic maps. The orange, red, and dark red distributions show a rise in traffic intensity at the beginning of the morning rush hour, reaching their maximum in the middle of the day, and then gradually decreasing as the evening rush hour begins.
At the \MER station, this pattern is more pronounced: traffic intensities in the red and dark red distributions continue to increase until 2~pm. In contrast, the other three stations exhibit flatter distributions after reaching their peaks by the end of the morning rush hour. This behaviour is consistent with the topologies of the areas considered. For instance, \MER is located near the city’s central distribution markets—which close early—and is also adjacent to a major road that crosses the city. In comparison, the others are situated in more suburban and residential areas, with less variations until the end of the day.

\subsection{Pollution and Traffic Correlations}
\label{subsec:pollutiontrafficcorr}
\begin{figure}[htb!]
	\begin{center}
		\includegraphics[width=0.45\linewidth]{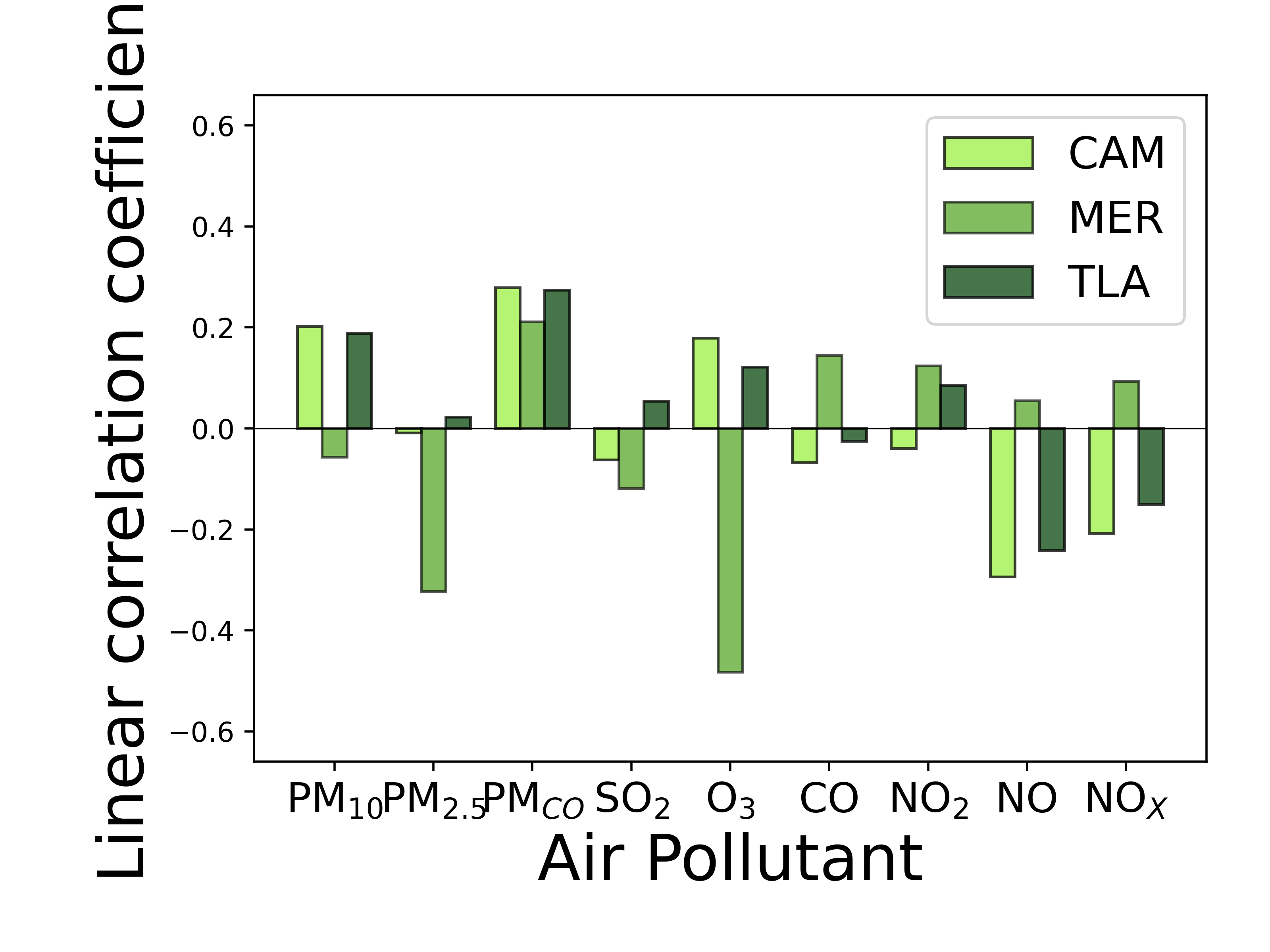}
		\includegraphics[width=0.45\linewidth]{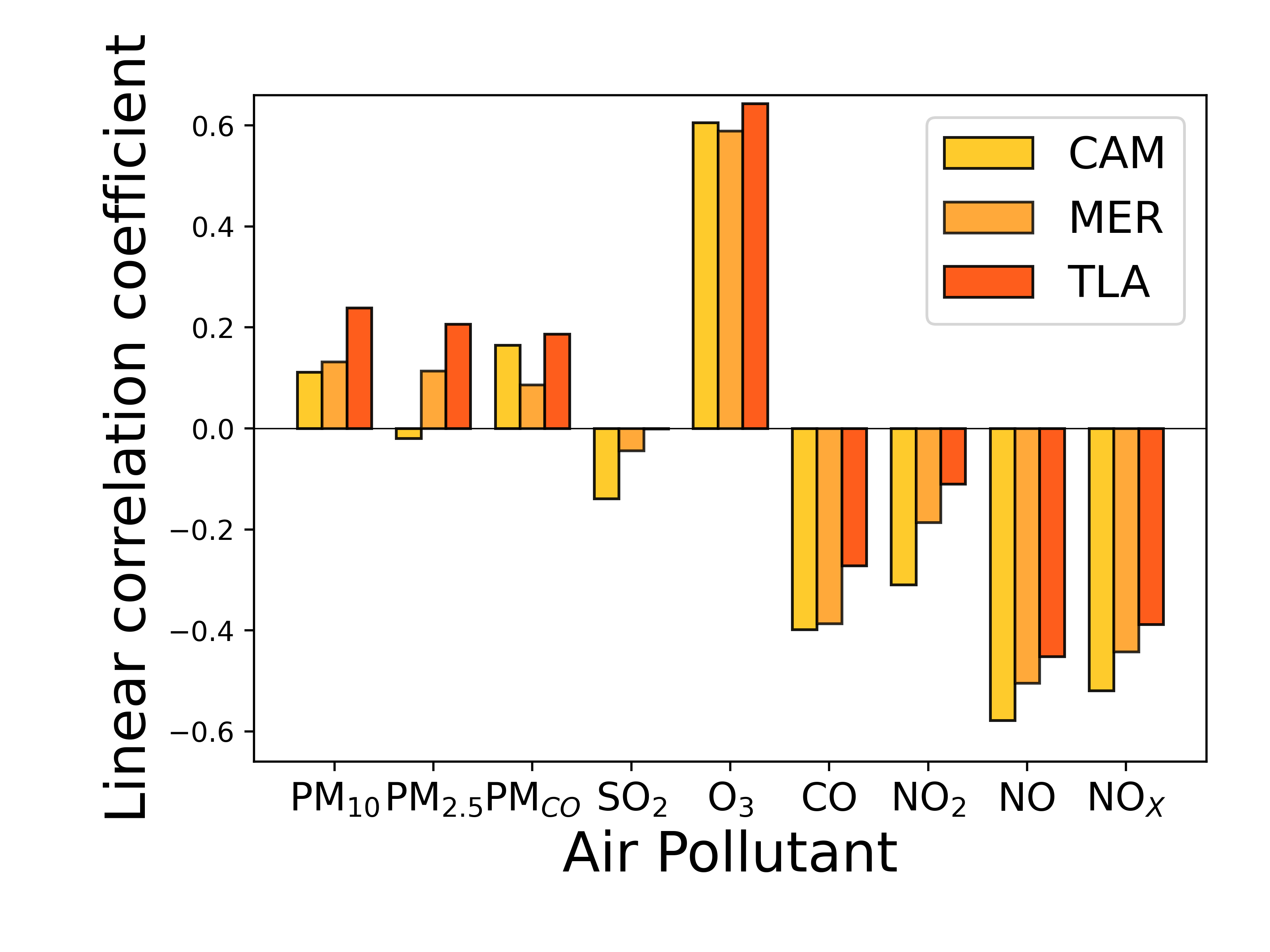}
		\includegraphics[width=0.45\linewidth]{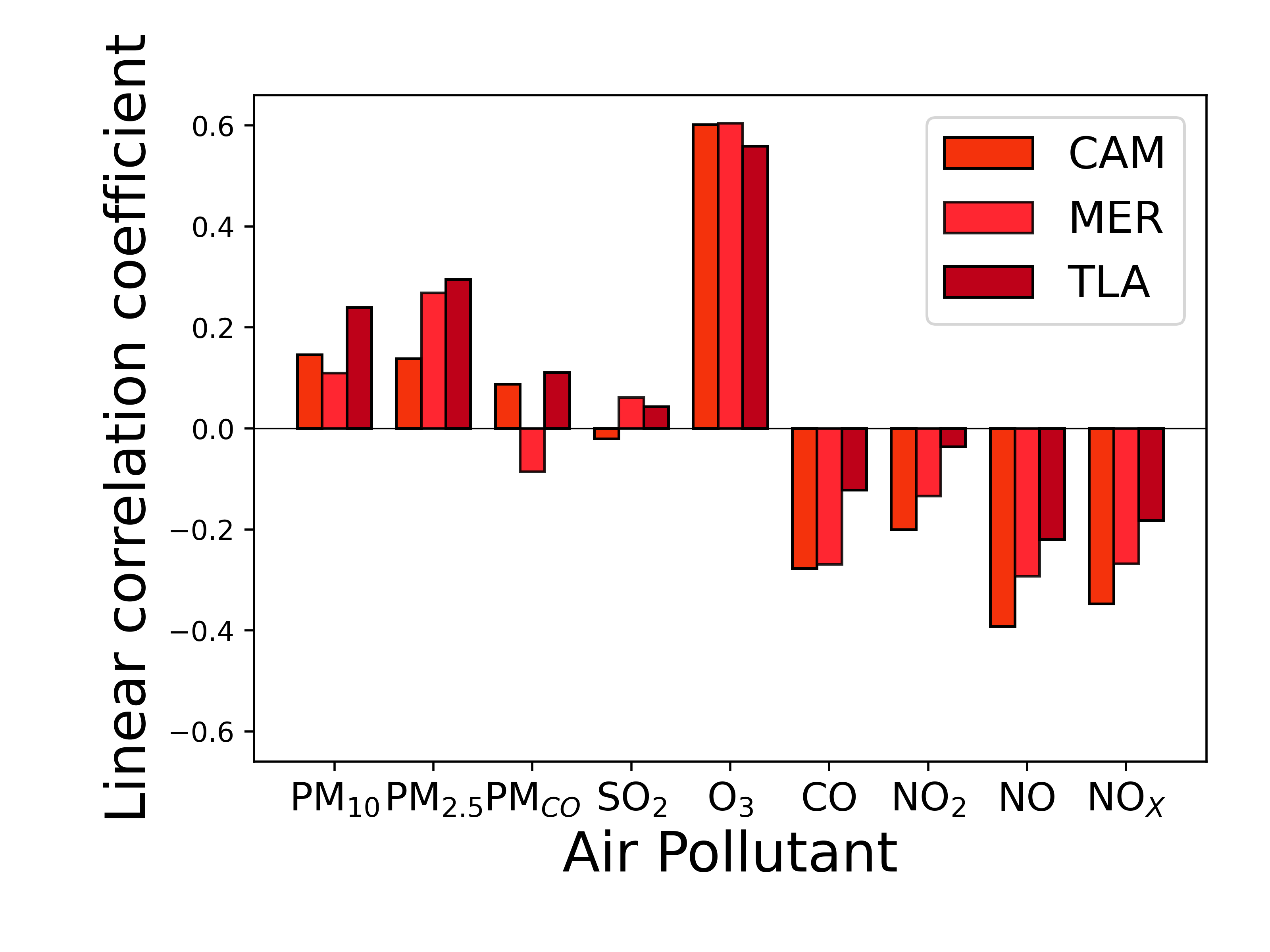}
		\includegraphics[width=0.45\linewidth]{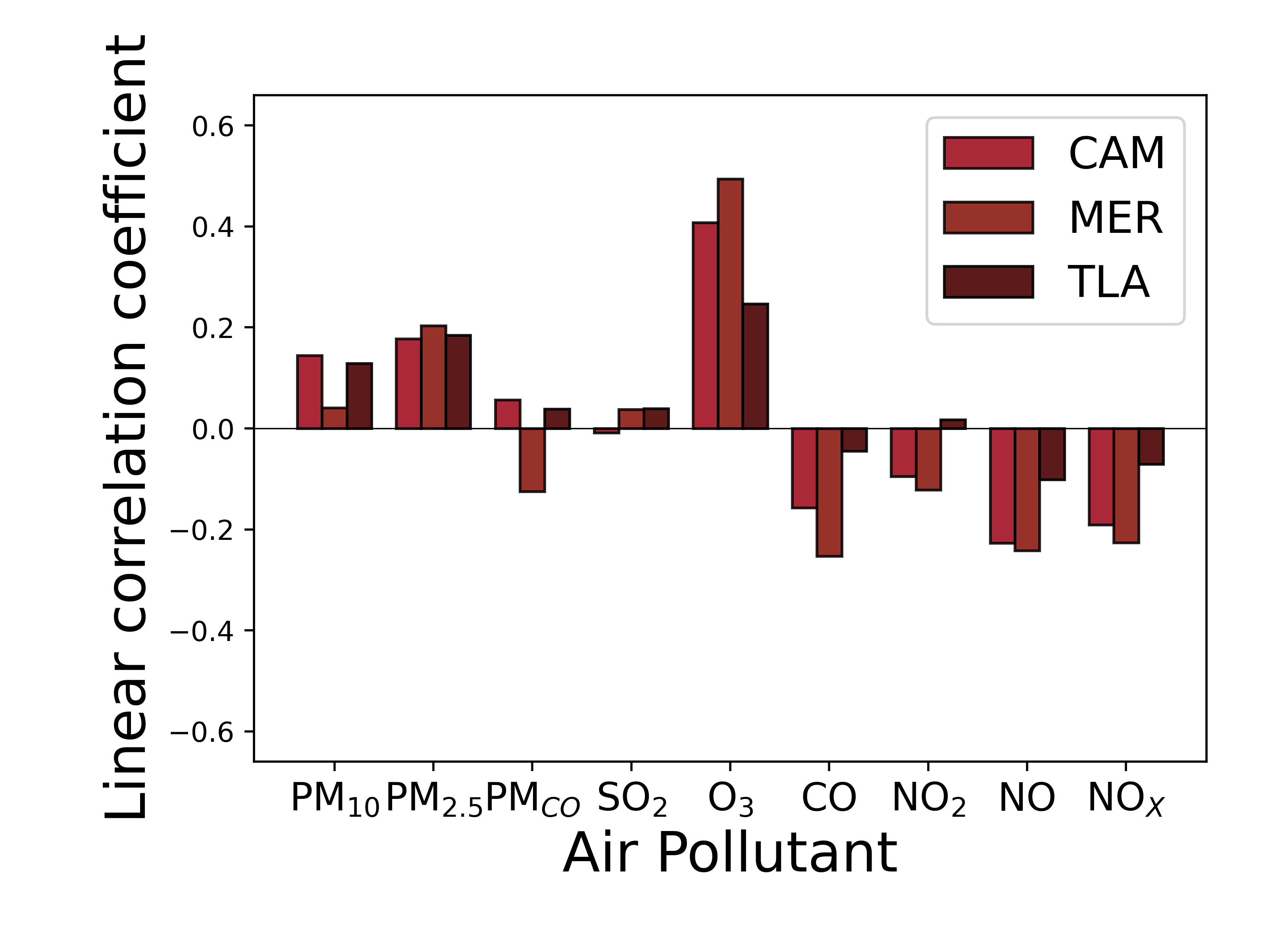}
		\caption{The linear correlation coefficients between the concentration of air pollutants and the traffic intensity, for each of the four traffic colours. This is shown for each of the three stations (\CAM, \MER, and \TLA).}
		\label{fig:CorrelationPollutants}
	\end{center}
\end{figure}

Fig.~\ref{fig:CorrelationPollutants} shows the linear correlations between the traffic intensities and the pollutants for the four traffic colours and the three benchmark stations considered (\CAM, \MER, and \TLA).

We note that considering orange, red and dark red traffic, \Othree shows a large positive correlation, while \CO, \NOtwo, \NO and \NOX all have negative correlations with these colours.
We see in Fig.~\ref{fig:pollutantshourly} that \Othree has a peak in the middle of the day, which corresponds to an increase in temperature and a depletion of other chemicals such as \NOtwo and \NOX as detailed in Sec.~\ref{subsec:PollutantsConcentration}.
The orange, red and dark red traffic intensities also peak in the middle of the day. However considering that also the temperature would peak in the middle of the day, it is not clear how much the correlation between \Othree and the high-traffic colours is driven by a specific effect from traffic on the pollutants or rather by a dominant correlation with the temperature.
The green traffic does not show a clear relation between its intensity and pollutants across the three stations and the particulate matter and \SOtwo pollutants do not show any strong correlations with the traffic colours.

\subsection{Analysis Workflow}
\label{subsec:analysisworkflow}

We summarise here the overall analysis workflow underlying each of the analyses and results shown in this paper. We start from the pollution and traffic data cleaned and preselected as described above and we follow the strategy below. 
\begin{enumerate}
  \setlength{\itemsep}{0cm}
  \setlength{\parskip}{0cm}
\item The pollution and traffic data used for PLSR development are split into 80\% for training and 20\% for testing.
\item The training sample is standardised by applying the z-score transformation to all variables.
\item Five-fold cross-validation is performed on the standardised training sample.
\item The average RMSE across the five training runs is compared for models with increasing numbers of components, and the optimal number of PLSR components is chosen as the one that minimises the RMSE.
\item The PLSR model is retrained on the whole training sample with the optimised number of components.
\item The PLSR model is applied on the test sample to assess the performance and check for overtraining
\item To assess compatibility, the distributions of all variables are compared between the training plus testing sample and the validation sample, ensuring their ranges are similar.
\item Finally we evaluate the model on the validation data checking the various metrics, the residuals and comparing measurements and predictions.
\end{enumerate}

We employed the Scikit-learn library (version 1.5.1) in Python to standardise the data, create the test-train split, and run the PLSR algorithm using the 5-fold cross validation~\cite{scikit-learn}.

\section{Data Modelling and Results}
\subsection{Partial Least Squares Regression Model}
\label{subsec:regression}
The PLSR may be considered as a mixture of principal components regression (which is based on
principal component analysis) and canonical-correlation analysis. It tries to find a linear regression model by projecting the predictor variables (organised in n observations-by-p predictors data matrix X) and the response variables
(organised in n observations-by-m responses data matrix Y ) to new hyperplanes or sub-spaces. These hyperplanes or sub-spaces are linear combinations of the original predictor variables (columns of X) and response variables (columns of Y ), respectively, and are determined by the canonical-correlation analysis to have maximum correlation with each other. In this subsection, we provide a brief summary on the PLSR technique. For a more complete exposition of this topic with more technical details, we refer the readers to Refs.~\cite{esposito2010handbook,rosipal2005overview,haenlein2004beginner}.

The underlying model of PLSR can be formulated as:
\begin{equation} \label{eq:PLSReqn}
\begin{aligned}
X = X_S X_L^\mathrm{T} + X_{\text{residuals}} \\
Y = Y_S Y_L^\mathrm{T} + Y_{\text{residuals}},
\end{aligned}
\end{equation}
where $X$ is an ($n$ observations)-by-($p$ predictors) data matrix, and $Y$ is an ($n$ observations)-by-($m$ responses) data matrix. Loading matrices $X_L$ and $Y_L$ are ($p$ predictors)-by-($n_{\text{comp}}$ PLSR components) and ($m$ responses)-by-($n_{\text{comp}}$ PLSR components), respectively. Predictor scores $X_S$ and response scores $Y_S$ are ($n$ observations)-by-($n_{\text{comp}}$ PLSR components) matrices, which are linear combinations of predictors in $X$ and responses in $Y$, respectively, and columns of $X_S$ are called PLSR components, having maximum covariance with columns of $Y_S$. Matrices $X_{\text{residuals}}$ and $Y_{\text{residuals}}$ are the residuals or error terms of the modelling, assumed to be independent and identically distributed random variables.

\begin{figure}[htb!]
	\begin{center}
		\includegraphics[width=0.75\linewidth]{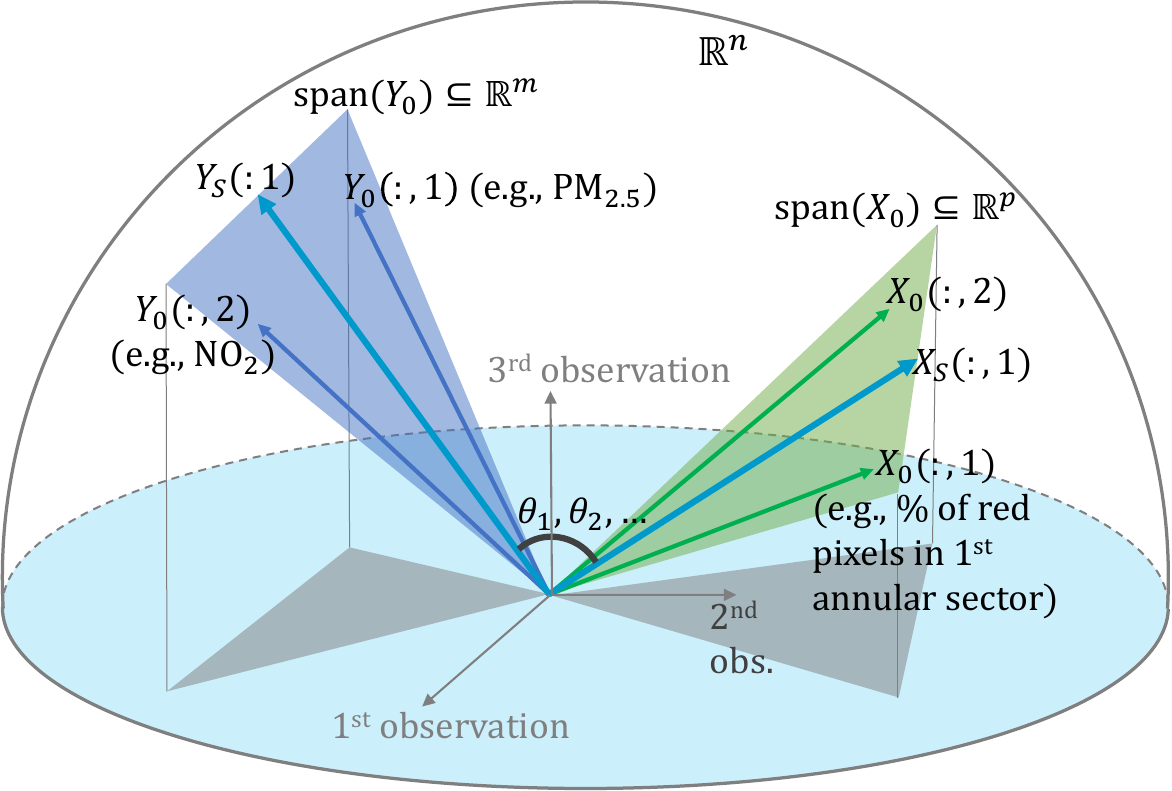}
 		\caption{Principal angles between flats or subspaces $\text{span}(X_0)$ and $\text{span}(Y_0)$. The $\text{span}(X_0)$ and $\text{span}(Y_0)$ are column spaces of centred and normalised traffic predictors $X_0$ and pollutants responses $Y_0$, respectively. We used MATLAB's syntax $X_0(:,1)$ and $Y_0(:,1)$ to denote the first columns of matrix $X_0$ and $Y_0$, respectively, and each column vector is $n$-dimensional containing $n$ observations. The columns of $X_S$ and $Y_S$ are linear combinations of columns of $X_0$ and $Y_0$, respectively, such that $\{ \cos{\theta_1},\cos{\theta_2},\ldots \}$ (\ie the canonical correlations) are maximised and sorted in non-increasing order.}
 		\label{fig:PrincipalAngles}
 	\end{center}
\end{figure}

In actual model fitting and regression, a ($p$ predictors)-by-($m$ responses) coefficients matrix $\beta_{n_{\text{comp}}}$ is fitted minimising the squares of the residuals according to
\begin{equation} \label{eq:PLSbeta}
\begin{aligned}
Y_S Y_L^\mathrm{T} = X_S X_L^\mathrm{T} \beta_{n_{\text{comp}}}, \\
\end{aligned}
\end{equation}
using $n_{\text{comp}}$ PLSR components, where the $n_{\text{comp}}$ can be fixed by cross-validation to minimise the RMSE in $Y_{\text{residuals}}^2 = (Y_0 - Y_S Y_L^\mathrm{T})^2$, and the corresponding $X_S X_L^\mathrm{T}$ and $Y_S Y_L^\mathrm{T}$ are ``reconstructions'' of the centred predictors and responses data matrices $X_0$ and $Y_0$ using $n_{\text{comp}}$ PLSR components, respectively. In this sense, the spatial dimension in the predictors matrix $X_0$ is effectively reduced to $n_{\text{comp}}$ in the fitted regression model, where the coefficients matrix $\beta_{n_{\text{comp}}}$ is effectively determined by $Y_S = X_S\beta$, resulting in a parsimonious model with optimal predictive power. 
This is visualised in the schematic in Fig.~\ref{fig:PrincipalAngles}, where we take the traffic predictors and pollutant responses as examples.

In our analysis, the PLSR was trained to predict the nine pollutant measurements from the 60 traffic intensities, hence we used $p=60$ traffic predictors and $m=9$ pollutant response variables from eq.~\ref{eq:PLSReqn}. 

We define a benchmark model as the model that was trained and tested on the data from three stations (\CAM, \MER and \TLA). We then validated this model using station \PED. 
Then the model was trained using six stations (\CAM, \MER, \TLA, \SAG, \SFE, and \UIZ), while still using \PED as the validation station.
Finally using the result of a station similarity analysis (see below in Sec.~\ref{subsec:stationsimilarity}), we selected one station as the “most similar" to the validation station \PED~to train an alternative model. For these three scenarios, the $n$ observations (i.e., the number of records used for training) vary from $711$ to $4093$.

Regarding the PLSR components, we used seven $n_{\text{comp}}$ for the three-station benchmark, four for the six-station model, and nine for the “most similar station" model.

\subsection{Station Similarity Analysis}
\label{subsec:stationsimilarity}

To test whether using similar stations for training and validation improves the model prediction, we define similarity as the stations having a similar distribution of traffic colour intensities in each of the 15 rings. To calculate a “similarity score" we consider the difference between the distribution of each colour in each ring, but weight these by how important it is in the benchmark model. For example, we want to weight dark red traffic as less important in the similarity score because it is more rare and has less impact on how the model performs.

For this scope we use the Variable Importance (VIP) Score and calculated as:
\begin{equation}
\textrm{VIP}_{j} = \sqrt{\frac{\sum w^{2}_{jf} \cdot \textrm{SSY}_{f} \cdot J}{\textrm{SSY}_\textrm{total}}} 
\label{eq:vipscore}
\end{equation}
where the index $j$ runs on the input variables, $w$ is the weight for the $j^{th}$ variable for PLSR component $f$, SSY is the sum of squares of variance explained in total and $J$ is the number of input variables $X$. SSY$_\textrm{total}$ is the total sum of squares of the variance explained of dependent variables $Y$~\cite{vip_score}.

\begin{figure}[hbt!]
	\begin{center}
        \includegraphics[width=0.65\textwidth]{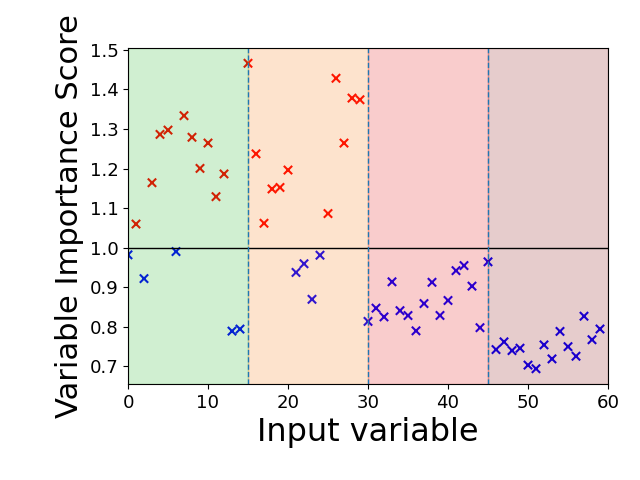}
        \vspace{-0.3cm}
        \caption{VIP scores of all the 60 input variables (15 rings for four colours), calculated based on the PLSR weights of the benchmark model using three stations for training.}
    \label{fig:VIP_scores}
    \end{center}
\end{figure}

Fig.~\ref{fig:VIP_scores} shows the Variable Importance in Projection (VIP) score (see Eq.~\ref{eq:vipscore} and Ref.~\cite{vip_score}), extracted from the benchmark model training and calculated for each of the 60 traffic predictors split by colour. A predictor with a score greater than one is considered to be important in the model as seen by the red markers on the plot.

Using this information, we then calculate a weighted chi-square score~\cite{chisquare}: 
\begin{equation}
\nonumber
\chi^2_\omega  = \sum_j\sum_k \left( \frac{(O_{jk}^{(u)} - E_{jk}^{(u)})^2}{E_{jk}^{(u)}} + \frac{(O_{jk}^{(v)} - E_{jk}^{(v)})^2}{E_{jk}^{(v)}} \right)\omega_j
\end{equation}

where index $k$ runs over the bins of the distribution of each $j$ input variable, $O_{jk}^{(u)}$, $O_{jk}^{(v)}$ are the observed frequencies in bin $k$ from group $v$ and $u$, 
$E_{jk}^{(u)}$, $E_{jk}^{(v)}$ are the expected frequencies in bin $k$ for each group, and
$\omega_j$ is the normalised VIP score for each input variable.
The expected frequencies are computed as:

\begin{equation}
\nonumber
E_{jk}^{(u)} = \frac{O_{jk}^{(u)} + O_{jk}^{(v)}}{N_u + N_v} \cdot N_u, \quad
E_{jk}^{(v)} = \frac{O_{jk}^{(u)} + O_{jk}^{(v)}}{N_u + N_v} \cdot N_v
\end{equation}

where $N_u$ and $N_v$ are the total counts in each group.

This compares the weighted $\chi^2$ score of the 60 input variables between two stations (the groups in the formalism above) and weight them by multiplying the VIP score, normalised to be between 0 and 1, to produce an overall similarity score for each pair of stations: the lower the score, the more similar the stations.
Using this metric, we identified the \SAG station as the most similar to our \PED validation station. We then use \SAG data to train an alternative model, comparing it to the other two models using three or six stations.

\subsection{Partial Least Square Regression Analysis Results}
\label{subsec:PLS_analysis_results}
\begin{figure}[hbt!]
	\begin{center}
        \includegraphics[width=1\textwidth]{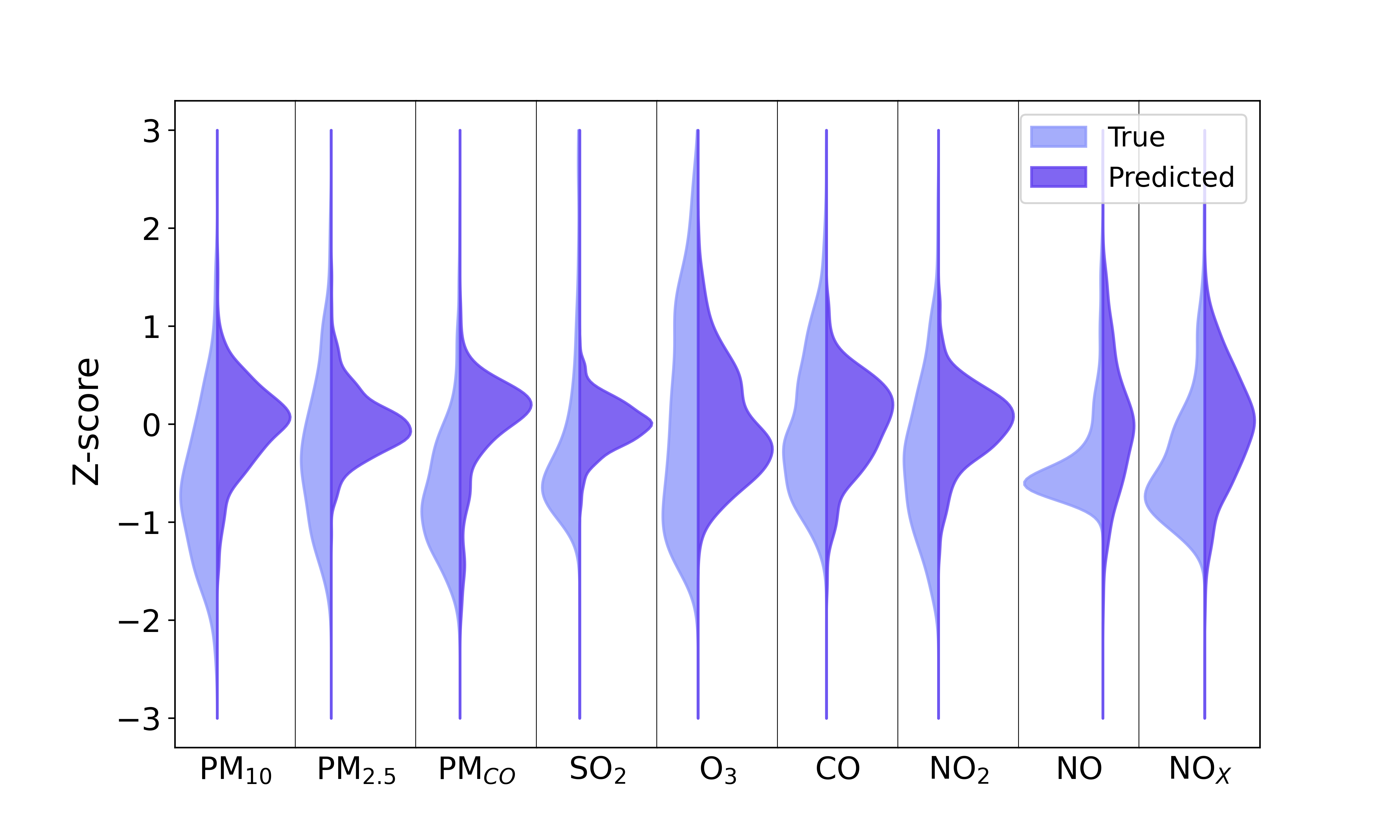}
        \vspace{-0.3cm}
        \caption{Distribution of standardised pollutant measurements comparing the true \PED distribution with the predictions from the benchmark PLSR model trained using three stations.}
    \label{fig:violin_truevspredicted-3_stations}
    \end{center}
\end{figure}

Fig.~\ref{fig:violin_truevspredicted-3_stations} shows the comparison between predicted and measured pollution levels in the validation station \PED using the benchmark training. The model is able to predict the distributions of the nine pollutants with different degrees of accuracy for the various pollutants. Good modelling is obtained for \Othree and \CO with well-behaved residual distributions centred at zero, the nitrous oxide pollutants show a less than one sigma bias in their residual distributions, while particulate matter pollutants and \SOtwo are less well predicted, with biases between one and two standard deviations and non-Gaussian residual distributions.

We note that the model predictions tend to have a narrower range compared to the true measurements. In particular, it can be seen when looking at the distributions of the particulate matter group of pollutants and \SOtwo. This may be because these pollutants have lower correlation with traffic than the other pollutants do, and the PLSR model is trained to optimise the predictions of the linear combinations of the nine pollutants (using the linear combinations of the traffic rings), instead of the distributions of specific individual pollutants. In addition, to avoid overtraining, the number of components retained in the PLSR is optimised and limited. Hence, pollutants with lower correlation to traffic will have their predictions focused on the value range captured by the correlations and the selected PLSR components.
In future iterations of this project we may want to revisit how we classify the traffic intensity by area, for example different aggregation of the traffic rings, and how we use the various pollutants in the model, as they have different correlations with the traffic intensities.

\begin{figure}[htb!]
	\begin{center}
    \hspace{-1.cm}
		\includegraphics[width=0.5\linewidth]{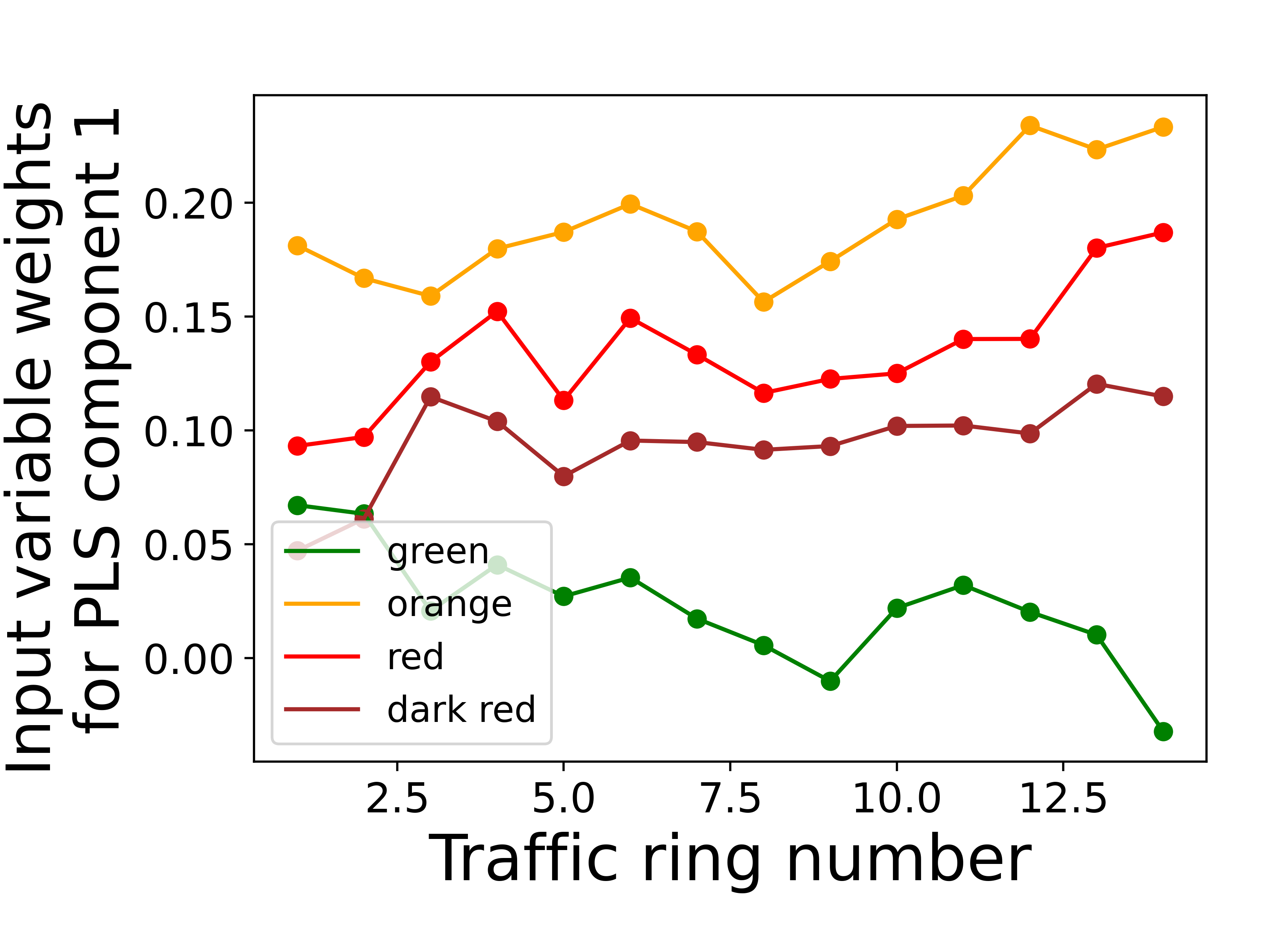}
        \hspace{-0.2cm}
  		\includegraphics[width=0.5\linewidth]{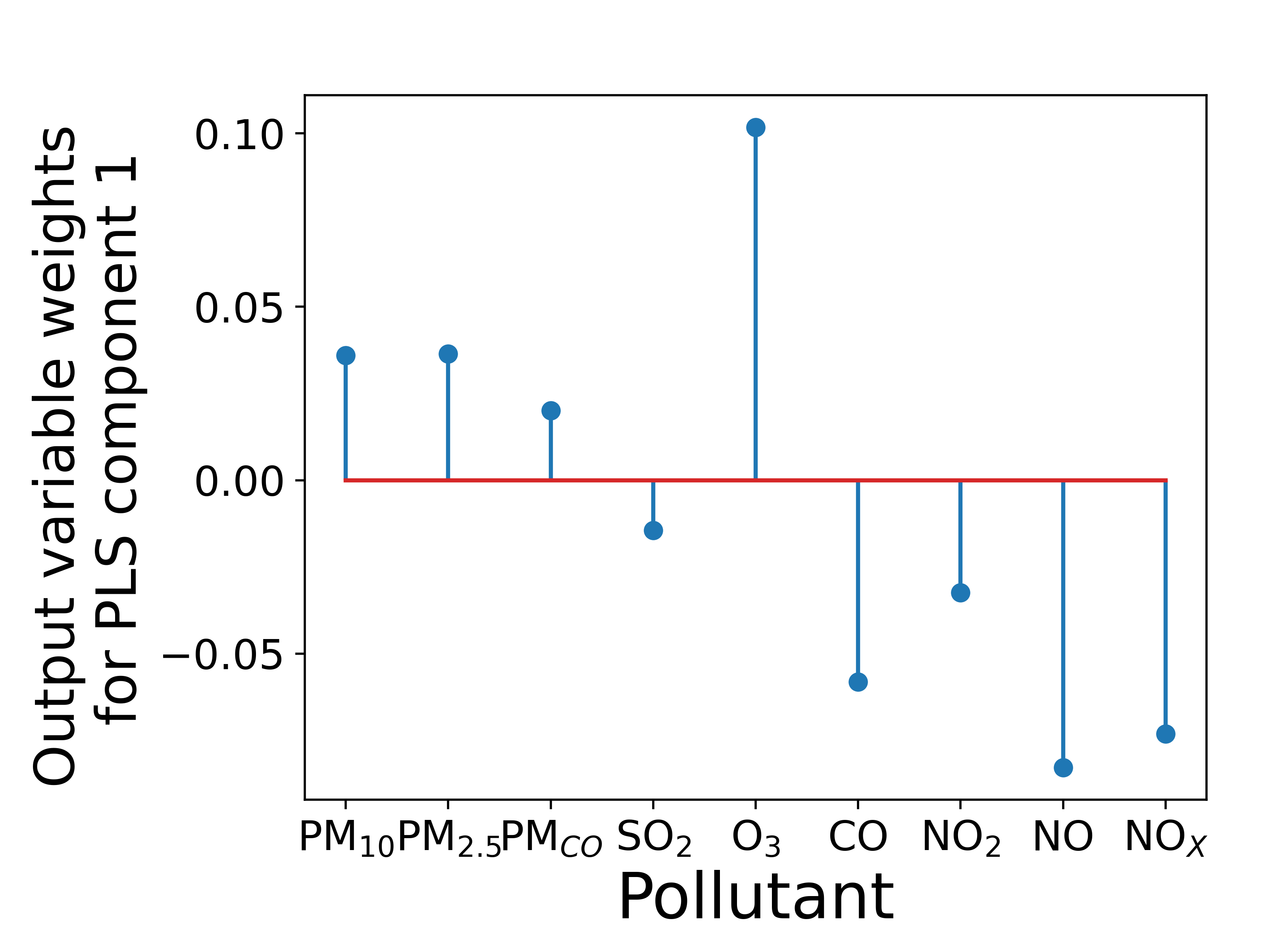}
        \hspace{-1.6cm}
        \caption{Input and output weights of the first PLSR component based on the benchmark model using three stations for training.}
        \label{fig:CAM_MoToFr_PLSComp1}
	\end{center}
\end{figure}

Fig.~\ref{fig:CAM_MoToFr_PLSComp1} shows the weights calculated for the first PLSR component for both the traffic input variables and the pollution output variables.
On the left plot, we can see that the orange coloured pixels contribute the most to predicted pollutant concentration. This can be explained by this type of traffic corresponding to a driver frequently speeding up and slowing down. On the right plot, we see a similar pattern as in Fig.~\ref{fig:CorrelationPollutants}, specifically for the orange traffic which dominates this component.

Looking again at the VIP scores in Fig.~\ref{fig:VIP_scores}, we see agreement with Fig.~\ref{fig:CAM_MoToFr_PLSComp1} as, overall, the green and orange traffic colours have a greater importance across their 15 rings and the highest score is obtained by the first orange ring. The red and dark red predictors have VIP scores of mostly less than one and this could justify removing those traffic colours to simplify the model in future iterations. 
It is hard to see a pattern across the rings from 1-15 within a given colour, as there is no clear trend.

\begin{figure}[hbt!]
	\begin{center}
            \includegraphics[width=1\textwidth]{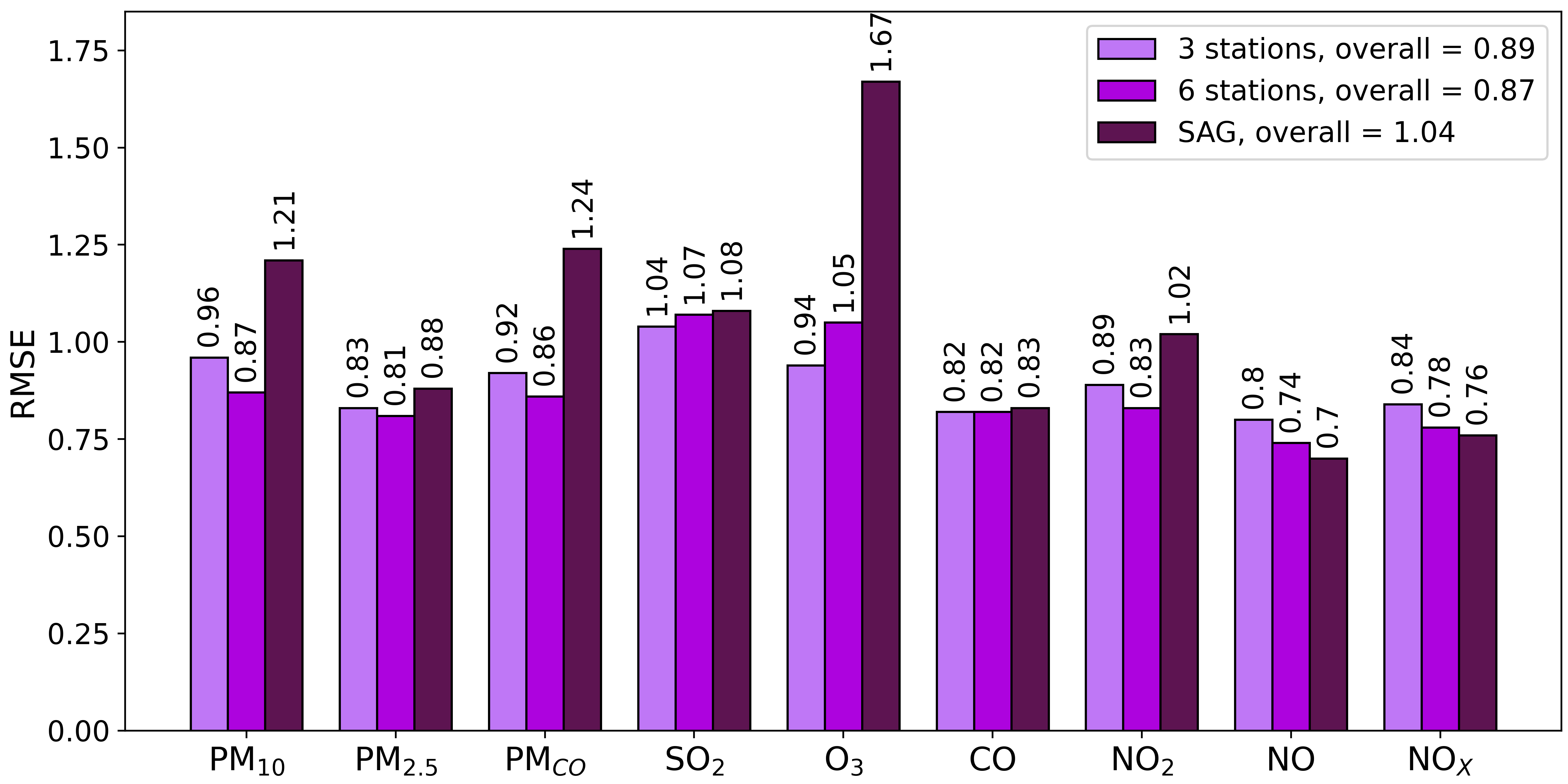}
	\end{center}
    \caption{Root Mean Squared Errors (RMSE) for each pollutant for the three PLSR training scenarios.}
    \label{fig:overall_results_RMSE}
\end{figure}

To improve the model, we looked at using more data by training on six stations to see whether including more information from different types of road layouts may help improving the predictions.
The overall Root Mean Squared Errors (RMSE) is lowest when six stations are used for training as shown in Fig.~\ref{fig:overall_results_RMSE}, while it is higher for the single \SAG station training pointing at the benefit of more data over station similarity.

\section{Conclusions}
\label{sec:DiscussionConclusions}
The main purpose of this paper is a proof of concept for quantifying traffic intensity starting from Google Maps traffic images, in order to predict air pollution using a simple linear model.
Current literature does not provide specific procedures on collection and treatment of traffic data, while the variability of the available sources is huge in both the type of measurements and their confidence levels.
We designed an innovative way to represent the traffic intensities from simple colour-coded traffic maps via concentric rings. This allows us to connect improved traffic descriptions to the pollution measurements available.

The concentric rings and the total traffic intensity defined represent a general way to characterise traffic levels that can be used starting from any source of data and that can be optimised according to the purpose of use. For example different ring segmentations can be used (equal radius vs equal area) and different weights can be calculated for each ring in the total intensity. The main idea is to characterise the traffic behaviour of a whole area rather than using, e.g., the single closest street segment as much of the existing literature is currently using. 

The purpose of the modelling in our project is two-fold: (1) to map the traffic activities to pollutant concentrations in order to reveal insight on their detailed relations, and (2) to develop a pollutant concentrations prediction system based on traffic information. Most black-box machine learning techniques are well competent in (2) only and not capable to provide interpretable insight for (1). Unlike those black-box techniques, the PLSR technique is competent in both (1) and (2), which is the main reason why it is employed in our analysis.

The PLSR allows us to obtain predicted pollutant distributions using traffic intensity distributions as input and developing a workflow that can be generalised to any source of traffic data and easily applied to any urban area. The pollution data we used are standard measurements provided by most city sensors; hence, the analysis can be readily applied to any other city and used to enhance existing sensor networks, which may be coarse or insufficiently reliable.

The PLSR modelling results can be interpreted as reflecting the correlation studies we performed on the traffic intensities and the pollutants. Good modelling is obtained for \Othree and \CO with well-behaved residual distributions centred at zero. These pollutants have clear correlations with traffic, especially with orange traffic levels. The nitrous oxide pollutants show also well-behaved residual distributions, but with a small effect of overestimation of their values (residual bias at the level of less than one standard deviation). The particulate matter and \SOtwo pollutants are the ones less well predicted, with bigger biases (between one and two standard deviations) and non-Gaussian residual distributions, being also the ones with lower correlation with traffic than the other pollutants.

As already noted, in general, the model predictions tend to have a narrower range compared to the true measurements. The PLSR model relies on maximising the correlation between the traffic set and the pollution set and the PLSR components taken are a smaller number than the input dimensions. Hence pollutants with lower correlation to traffic will have their predictions focused on the value range captured by the correlations and the selected PLSR components.

The results of the PLSR model indicate that error is reduced more effectively by increasing the amount of training data than by using traffic inputs more similar to those of the validation station. This suggests that incorporating data from six stations would capture a broader range of traffic and pollution information, thereby enhancing predictive accuracy.
However, using the six stations from this three-month period still gives us a relatively small increase in the training dataset ($4093$ records) with respect to the overall size of the feature space generated by the $60$ predictors. In future work we will be able to exploit a much larger time period to give the model a significantly larger coverage of the input feature space and we will study an appropriate reduction of the number of rings, i.e., the number of predictors.

The benefit of using PLSR to model the pollutants is that it is very simple to implement and it is suited for datasets with high collinearity. Other machine learning algorithms may be used in the future to improve prediction power as they are able to better model non-linear relationships between input and output features. Including the time of day and the weather information as extra input features could also benefit the predictive power of the model. 

Finally, we would like to underline that while “green" traffic is seen as a positive situation by drivers, in the context of our analysis it needs to be considered similarly to the other colours as it could be as detrimental to air quality as red and dark red traffic. In general, the various traffic levels are expected to correspond to different polluting profiles. For example, green traffic could sometimes correspond to vehicles travelling at higher speeds, hence this would lead to higher levels of \NOX~\cite{OptimalSpeedRanges, EmissionsFromVehicles}. On the contrary, for red traffic levels, one can expect lower travelling speed and a higher frequency of braking. The latter could lead to more tyre/break wear and hence to more particulate matter emission.

\subsection{Data and Code Availability}
\label{subsec:datacode}
The data generated in this study together with the source code that enables all experiments to be reproduced is available on Zenodo at Ref.\cite{ZenodoSAPIENS}.

\subsection{Acknowledgements}
\label{subsec:Acknowledgements}

We thank our institutions IPN and QMUL and in particular Professor Teresa Alonso, and Dr Sharon Ellis from QMUL who organised and led the Sandpit event in IPN to seed multidisciplinary research projects on the theme of Smart Cities. The original idea for the SAPIENS project comes from the SAPIENS group formed at this Sandpit and composed of Marcella Bona, Adriana Lara, Alberto Luviano-Ju\'arez and John Moriarty.
We are grateful for the continuous support from IPN and QMUL that have given us the possibility to explore this avenue of research and have provided us with IT facilities and invaluable experts (Daohai Li and Alex Owen in QMUL).

\subsection{Author contributions}


Conceptualisation: M.B., A.L., A.L.J.
Data Curation: J.-C.H., V.L.-S., F.M-G., J.R., N.V.
Formal Analysis: M.B, N.H, J.-C.H, J.R. 
Funding Acquisition: M.B., A.L.
Investigation: M.B, N.H, J.-C.H, J.R.
Methodology: M.B, J.-C.H.
Project Administration: M.B, A.L.
Resources: F.M-G., N.V.
Software: N.H, J.-C.H., J.R., X.S.Z
Supervision: M.B., A.L.
Validation: M.B, N.H, J.-C.H, J.R. 
Visualisation: N.H, V.L.-S., J.R., X.S.Z
Writing – Original Draft Preparation: M.B., A.L., J.-C.H., J.R.
Writing – Review \& Editing: M.B., J.R., V.L.-S.

\vspace{0.5\baselineskip}
\noindent
{\bf{Correspondence}} and requests should be addressed to Jocelyn Richardson at \href{mailto:jocelyn.richardson@qmul.ac.uk}{jocelyn.richardson@qmul.ac.uk} or Marcella Bona at  \href{mailto:m.bona@qmul.ac.uk}{m.bona@qmul.ac.uk}.

\subsection{Funding} 

The SAPIENS project started from seeding funding from QMUL and IPN under the “Global Research Collaboration Initiative", 2020. A.L. has received funding from the SIP-IPN under grant agreement No. 20251128. M.B., N.H. and J.-C.H. and X.S.Z received funding by the Science and Technology Facilities Council (STFC) Impact Acceleration Account. J.R. receives funding from Queen Mary University of London via a Principal's PhD Studentship. V.L.-S. receives financial support from the Secretaría de Ciencia, Humanidades, Tecnología e Innovación (SECIHTI) under grant number 960525; the Fundaci\'on Polit\'ecnico and Queen Mary University of London through the dual doctoral programme between the IPN and QMUL; and the Comisi\'on de Operaci\'on y Fomento de Actividades Acad\'emicas (COFAA-IPN).

\subsection{Competing interests}

The authors declare no potential conflict of interest. The funders had no role in the design of the study; in the collection, analyses, or interpretation of data; in the writing of the manuscript; or in the decision to publish the results.

\bibliographystyle{ieeetr}
\bibliography{SAPIENS} 

\end{document}